\documentclass[runningheads]{llncs}
\usepackage[final,year=2026]{eccv}
\usepackage[table, dvipsnames]{xcolor}
\usepackage{amsfonts}
\usepackage{amsmath}
\usepackage{bbm}
\usepackage{bm}
\usepackage{booktabs}
\usepackage{caption}
\usepackage{diagbox}
\usepackage{eccvabbrv}
\usepackage{graphicx}
\usepackage{multirow}
\usepackage{wrapfig}
\usepackage{xspace}

\usepackage[accsupp]{axessibility}
\usepackage[pagebackref,hidelinks]{hyperref}
\usepackage{orcidlink}

\definecolor{our_green}{rgb}{0.93, 0.98, 0.9}

\newcommand{\name}{MeshReGen\xspace}
\newcommand{\vecset}{VecSet\xspace}

\makeatletter
\renewcommand{\paragraph}{%
  \@startsection{paragraph}{4}%
    {\z@}{-0.5em}{-0.5em}%
    {\normalfont\normalsize\itshape}%
}
\makeatother

\makeatletter
\renewcommand\subsubsection{\@startsection{subsubsection}{3}{\z@}%
                       {-8\p@ \@plus -4\p@ \@minus -4\p@}%
                       {-0.5em \@plus -0.22em \@minus -0.1em}%
                       {\normalfont\normalsize\bfseries\boldmath}}
\makeatother

\newcommand{\xb}{{\boldsymbol{x}}}
\newcommand{\yb}{{\boldsymbol{y}}}
\newcommand{\zb}{{\boldsymbol{z}}}

\newcommand{\z}{{\boldsymbol{z}}}

\newcommand{\Dc}{{\mathcal{D}}}
\newcommand{\Ec}{{\mathcal{E}}}
\newcommand{\Pc}{{\mathcal{P}}}

\newcommand{\Rb}{{\mathbb{R}}}

\begin{document}

\title{\name: A Unified 3D Geometry Regeneration Framework} 
\titlerunning{\name: A Unified 3D Geometry Regeneration Framework}

\author{Geon Yeong Park\thanks{Work done during an internship at Meta Reality Labs.}\inst{1,2} \and
Roman Shapovalov\inst{2} \and
Rakesh Ranjan\inst{2} \and
Jong Chul Ye\inst{1} \and
Andrea Vedaldi\inst{2} \and
Thu Nguyen-Phuoc\inst{2}}

\authorrunning{G.~Y.~Park et al.}

\institute{KAIST \and
Meta Reality Labs}

\maketitle
\begin{center}
\centering
\vspace{-0.3cm}
{\small\href{https://geonyeong-park.github.io/meshregen/}{https://geonyeong-park.github.io/meshregen/}}\\[0.2cm]
\captionsetup{type=figure}
\includegraphics[width=\linewidth]{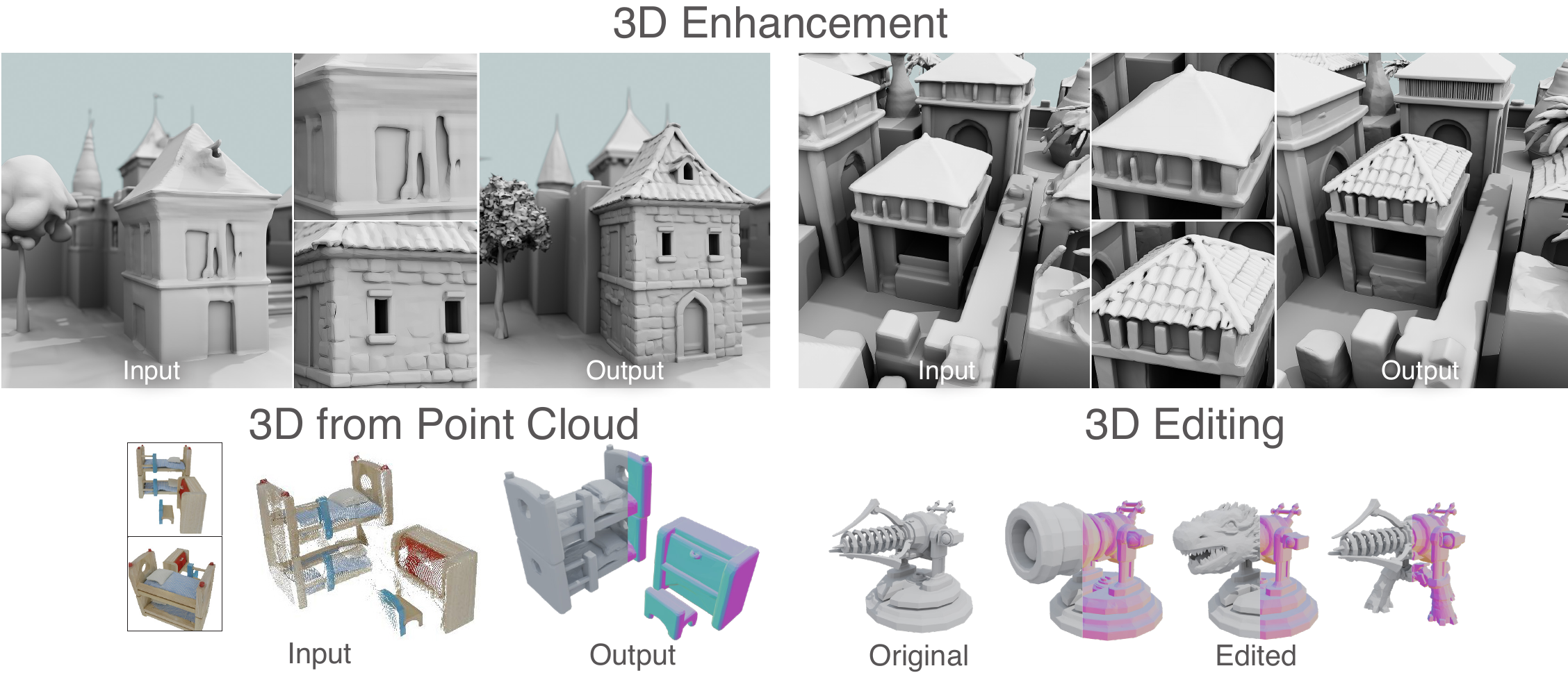}
\captionof{figure}{\textbf{\name} is a diffusion-based 3D regeneration framework that reconstructs complete 3D shapes from coarse geometry using 2D image cues.
The regeneration prior is learned through self-supervised pretext tasks and augmentations, without task-specific architectures or extra annotations.
Users can input a coarse mesh (\textit{3D Enhancement}), incomplete point cloud (\textit{3D from Point Cloud}), or masked mesh (\textit{3D Editing}) to recover detailed, complete 3D shapes.
}%
\label{fig:teaser}
\vspace{-0.4cm}
\end{center}%

\begin{abstract}

We consider the problem of \emph{regenerating} 3D objects from 2D images and initial 3D shapes.
Most 3D generators operate in a one-shot fashion, converting text or images to a 3D object with limited controllability.
We introduce instead \name, a 3D \textbf{re}generator that is conditioned on an initial 3D shape.
This conceptually simple formulation allows us to support numerous useful tasks, including 3D enhancement, reconstruction, and editing.
\name uses a new conditioning mechanism based on \vecset, which allows the regenerator to update or improve the input geometry with consistent fine-grained details.
\name learns a widely applicable regeneration prior from off-the-shelf 3D datasets via self-supervised pretext tasks and augmentations, without additional annotations.
We evaluate both the geometric consistency and fine-grained quality of \name, achieving state-of-the-art performance in controllable 3D generation across several tasks.
\vspace{-0.5cm}
\end{abstract}

\begin{figure*}[ht]
\centering
\includegraphics[width=\linewidth]{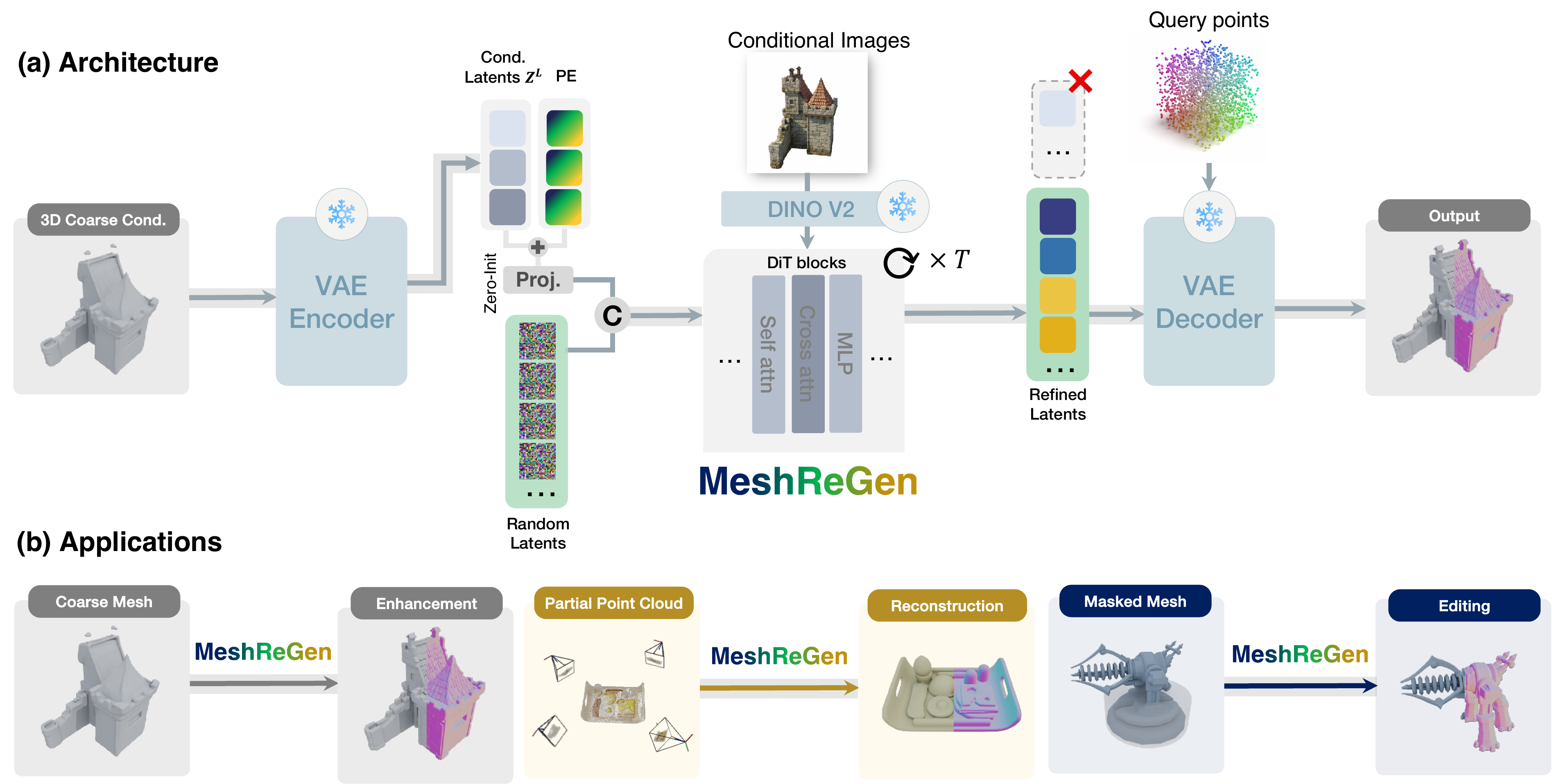}
\vspace{-0.7cm}
\caption{\textbf{Overview}. \name takes both 2D image and initial 3D geometry as input, enabling explicit control over global geometry (e.g., pose, coarse shape) while improving fine-grained details. The 3D condition is encoded as \vecset latents ($\zb^L$) that compactly represent global geometry. After summing with positional embeddings, these conditionings and random latents are diffused by a DiT into enhanced latents, then decoded into a complete high-quality 3D shape. 
}%
\label{fig:model}
\vspace{-0.6cm}
\end{figure*}

\section{Introduction}%
\label{sec:intro}

While 3D content creation is a complex process that requires significant expertise and effort, generative AI has the potential to radically transform it.
Similar to how language models have revolutionized coding, AI can accelerate 3D artist workflows and open 3D content creation to a much wider audience.
However, whereas coding is concerned primarily with a single modality, namely text, 3D content creation is a complex task that involves multiple tasks and modalities.

For example, 3D artists often start with a 2D concept art that captures the 3D content that must be created.
While the concept art can be captured by an image, the artist is often given additional specifications of a different nature.
An example workflow may have the artist sketch the 3D shape using a 3D sculpting tool, followed by iterative refinement or editing.
In other cases, the artist starts from a noisy scan (3D point cloud) of a real-life object, obtained using photogrammetry, and reconstructs that into a clean, faithful 3D shape that can be used in production.
All these examples require the 3D generation process to be controlled by 3D signals of various types (sculpts, scans, block-outs) in addition to 2D signals (concept art).

Recent advances in 3D generation~\cite{zhang233dshape2vecset:,xiang24structured,zhang24clay:,tripo3d24text-to-3d} have focused on an image-based workflow, where a 3D object $\xb^H \subset \Rb^3$ (a surface) is generated from a 2D image $I \in \Rb^D$.%
\footnote{This is true for text-based generation as well, as the text prompt is converted to a 2D image via a text-to-image model.}
Hence, generation amounts to drawing a sample $\xb^H$ from the distribution $p(\xb^H \mid I)$.
However, in the more elaborate workflows discussed above, it is necessary to control the generation process by an additional signal $\yb^L$, like a sculpt, a block-out, or a 3D scan.
While the nature and interpretation of these signals are widely different, there are some important similarities too.
First, all such 3D signals can be represented as 3D shapes $\yb^L \subset \Rb^3$.
Secondly, these 3D shapes $\yb^L$ contain \emph{significantly less (or partially distorted) information} than the target shape $\xb^H$.
For example, $\mathbf{y}^L$ may be (i) a coarse \textit{but} geometrically aligned proxy (e.g., voxel grid or bounding box) or (ii) a degraded and potentially misaligned geometry (e.g., a noisy scan or a distorted intermediate generation).
Hence, the common goal is to map a \emph{L}ow-information 3D shape $\yb^L$ to a \emph{H}igh-information 3D shape $\xb^H$.
In this paper, we thus ask whether, despite their different nature, all these tasks can be solved by a \emph{single} architecture and learning formulation.

We call this common problem 3D shape \emph{\textbf{re}generation}.
Formally, the goal of 3D regeneration is to draw a 3D shape $\xb^H$ from a distribution $p(\xb^H \mid \yb^L, I)$ where $I$ is a 2D image and $\yb^L$ is an auxiliary 3D shape that may contain significantly less information than $\xb^H$. Two key challenges arise in training general 3D regenerators: (a) limited understanding of task-agnostic \textit{architectural} design, and (b) the scarcity of paired 3D training \textit{datasets}. 

\paragraph{Architecture.}
Prior work has addressed specific instances of regeneration problems (e.g., editing~\cite{li25voxhammer:}, enhancement~\cite{deng24detailgen3d:}) using tailored task-specific architectures, whereas we seek a \textit{unified architecture} that can be applied to all types of 3D downstream regeneration tasks.
To design such a general-purpose architecture, we start from \vecset~\cite{zhang233dshape2vecset:}, a powerful and highly flexible 3D latent space representation.
\vecset is based on a pair of encoder and decoder functions $\zb = \mathcal{E}(\xb)$ and $\xb = \mathcal{D}(\zb)$, where $\zb \in \Rb^{N \times d}$ is a set of $N$ latent vectors of dimension $d$.
In prior works, this latent space has been used to train latent conditional diffusion models, usually in the form of a Diffusion Transformer~(DiT)~\cite{peebles23scalable} implementing a denoising function $\zb' = \Phi(\zb, I)$, where $\zb$ is a noisy latent and $I$ is a conditioning image.
As we show in \cref{sec:method,fig:model}, we can adapt this architecture to solve the 3D regeneration problem by conditioning the diffusion process on the input coarse shape $\yb^L$, also encoded in the same latent space $\zb^L = \mathcal{E}(\yb^L)$.
Rather than injecting 3D controls via a separate cross-attention pathway \cite{zhang24clay:, hunyuan3d2025hunyuan3d}, we encode both the conditioning shape $\yb^L$ and the target shape $\xb^H$ using the same VecSet encoder and concatenate their tokens as joint input to the DiT. This design minimally modifies the pretrained generator, allowing the 3D condition to be processed by the same self-attention mechanism used during pretraining. While cross-attention is effective for sparse and simple controls (e.g., bounding boxes), in general regeneration scenarios involving dense or noisy inputs (e.g., corrupted scans), we show that our simple \vecset-level concatenation strategy empirically outperforms cross-attention and other conditioning mechanism alternatives.

\paragraph{Dataset.} While recent efforts have explored 3D conditioning in generative models (e.g., CLAY~\cite{zhang24clay:},~\cite{hunyuan3d2025hunyuan3d}), their applicability to broader downstream tasks (e.g., enhancement, editing) remains unclear, partly due to the lack of scalable task-specific training \textit{dataset}.
Specifically, these methods are largely restricted to conditional generation given clean, geometrically accurate controls aligned with the target shape (e.g., bounding boxes).
Thus, it remains unexplored whether a single architecture can robustly handle the wider spectrum of 3D inputs $\yb^L$ with severe degradations or mismatched geometry (e.g., noisy scans, lossy-compressed assets, or distorted outputs from low-quality generators), limiting the range of inverse problems such priors can address. Although large-scale 3D base datasets are now generally available~\cite{deitke23objaverse-xl:}, these datasets lack the annotations required to learn 3D regeneration models (namely pairs $(\xb^H,\yb^L)$).
Another contribution of this paper is to introduce protocols that can automatically convert a generic 3D dataset into training data for 3D regeneration, where $\yb^L$ is a degraded version of $\xb^H$, obtained via data augmentation, without additional expensive annotations.

To show the effectiveness of our approach, we experiment in particular with three different tasks:
(\textbf{a})~Enhancement: Given a very low-resolution or even corrupted 3D component of a large 3D scene, we upgrade its quality;
(\textbf{b})~Editing: Given an initial 3D shape, we allow the model to change significant parts of it;
(\textbf{c})~Reconstruction: Given an initial 3D scan reconstructed from a small number of views using photogrammetry, we recover a full and geometrically faithful 3D object.
Importantly, all these tasks are solved using the same exact model architecture.
Our experiments show that \name achieves state-of-the-art results even compared to models that are designed specifically for each task.

\vspace{-0.25cm}
\section{Related Work}%
\label{sec:related-work}
\vspace{-0.25cm}

\textbf{Advances in 3D generative models.}
The increasing availability of large-scale 3D datasets~\cite{deitke23objaverse-xl:, deitke23objaverse:} has enabled rapid progress in foundation 3D generative models.
Some methods generate 3D assets using only 2D supervision from multi-view rendered images~\cite{siddiqui24meta, xu24instantmesh:}.
These approaches typically learn to generate multi-view images of the same object, which are used to reconstruct through 3D volumetric representations such as radiance fields, tri-plane features, or Gaussian splats, from which a 3D mesh can be extracted using Marching Cubes~\cite{lorensen87marching}.
However, the lack of explicit volumetric constraints can result in open surfaces or inaccurate interiors.
In contrast, methods with direct 3D supervision~\cite{li2024craftsman, xiang2024structured, li2025sparc3d, tripo3d24text-to-3d, lei25hunyuan3d, zhao2025hunyuan3d, zhao23michelangelo:, chen24dora:} can explicitly generate SDFs, from which 3D meshes can be cleanly extracted with arbitrary topology.

Most 3D generators take text or image prompts as input, which are intuitive but inherently under-constrained.
To improve controllability, recent works introduce explicit 3D-aware controls into native 3D generators, such as voxels, bounding boxes, or other structured primitives (e.g., CLAY~\cite{zhang24clay:}, PoseMaster~\cite{yan2025posemaster}, Hunyuan3D-Omni~\cite{hunyuan3d2025hunyuan3d}).

\textbf{3D generative priors for downstream tasks.}
Despite advances in 3D \textit{generation}, a broader question remains: how can 3D generative priors provide a unified framework across multiple \textit{downstream tasks}?
In 2D, foundation generative models increasingly support editing, reconstruction, and other tasks within a single architecture~\cite{wu2025qwen,yin2025qwen,labs25flux.1,team2023gemini}.
In contrast, 3D downstream applications often rely on task-specific architectures or training-free adaptations~\cite{deng24detailgen3d:,li25voxhammer:}.
A principled design for unified 3D task generalization and scalable construction of paired training data remains underexplored.
In this work, we take an initial step by addressing both architectural and data aspects, with the goal of encouraging progress toward unified, generalizable, and versatile 3D regeneration priors.

\section{Method}%
\label{sec:method}

\begin{figure*}[ht]
\centering
\includegraphics[width=\linewidth]{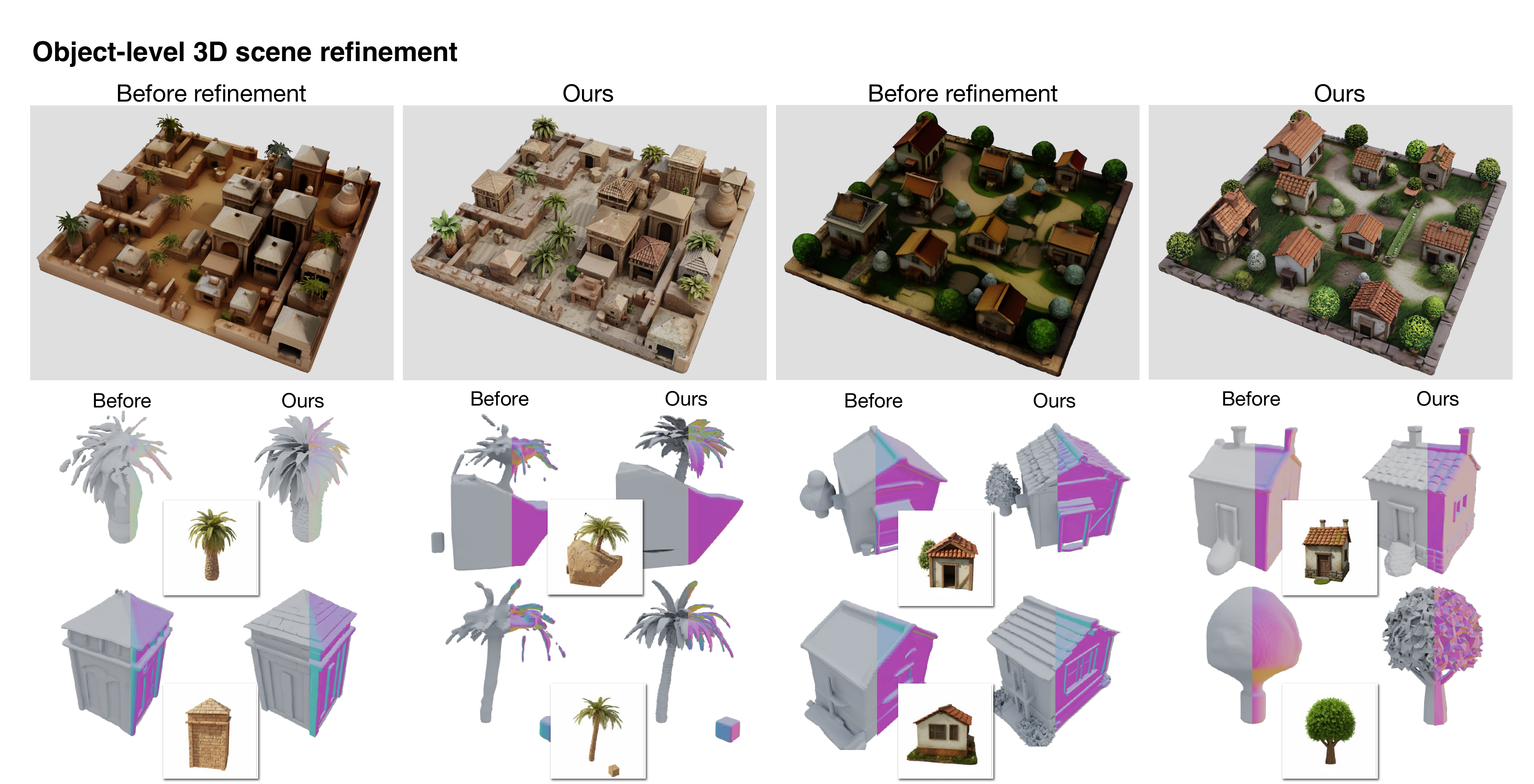}
\vspace{-0.5cm}
\caption{Qualitative results of object-level 3D scene refinement.
(Top) A single-scene image is converted into a coarse monolithic 3D scene by Sparc3D~\cite{li2025sparc3d}, where details of each object are degraded due to the limited capacity of the latent code.
To mitigate this degradation, we segment the scene with AutoPartGen~\cite{chen2025autopartgen}, refine individual objects with \name (Bottom), and update each object to produce a high-quality compositional 3D scene.
Image conditions are automatically regenerated from initial renders of each coarse object, as detailed in~\cref{sec:results_enhancement}, using an LLM-VLM\@.
Each scene and object are textured using a proprietary texture generator for visualization only.}%
\label{fig:scene_refine}
\end{figure*}
\vspace{-0.4cm}

\begin{figure*}[ht]
\centering
\includegraphics[width=\linewidth]{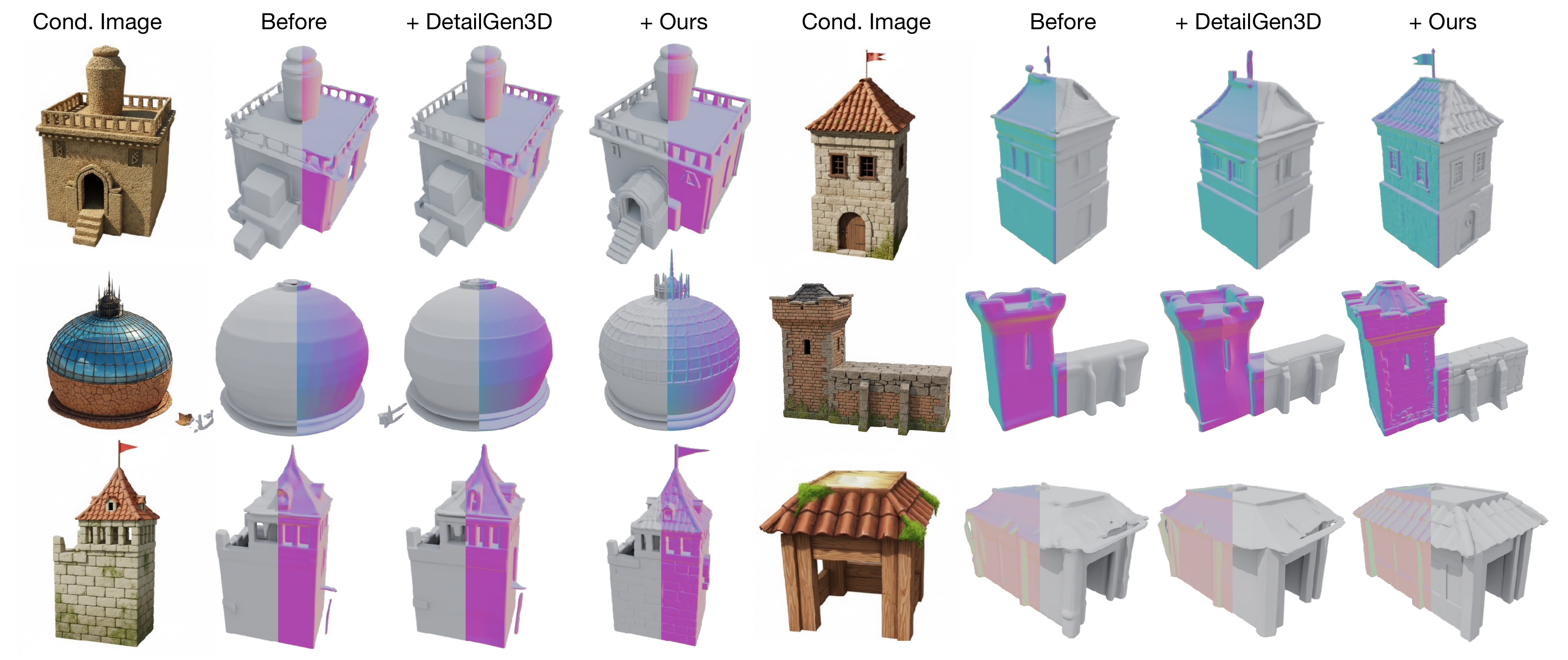}
\caption{
Qualitative comparison with 3D generation and enhancement baselines.
Coarse 3D conditions are obtained as parts of the original scene~\cite{li25sparc3d:}, and 2D image conditions are automatically regenerated given initial renders of each coarse object, as detailed in~\cref{sec:results_enhancement}.
Using these multi-modal conditions, \name regenerates the fine-grained details of 3D shapes, outperforming the recent enhancement baseline~\cite{deng24detailgen3d:}.
}%
\label{fig:obj_refine}
\vspace{-0.3cm}
\end{figure*}

Let $\xb^H \subset \Rb^3$ denote the surface of a 3D object that we wish to generate, where the suffix stands for \emph{H}igh information.
In \emph{3D regeneration}, we are interested in sampling $\xb^H$ from a probability distribution $p(\xb^H\mid \yb^L, I)$ conditioned on a 3D shape $\yb^L \subset \Rb^3$ (``\emph{L}ow information'') and, optionally, an image $I \in \Rb^D$.

\subsection{Model architecture}%
\label{sec:model-architecture}

\name builds upon a pre-trained latent diffusion model, operating in a compact latent space of 3D shapes~\cite{li25triposg:,li24craftsman:,zhao25hunyuan3d,xiang24structured,li2025sparc3d}, for which we use \vecset~\cite{zhang233dshape2vecset:}.

\subsubsection{3D Generation using \vecset.}%
\label{sec:vecset-latents-and-generator}

\vecset's~\cite{zhang233dshape2vecset:} encoder is a transformer neural network that maps a (centered) 3D shape $\xb \subset [-1,1]^3$ to $K$ $D$-dimensional latent vectors
$
\zb = \Ec(\Pc;K)\in \mathbb{R}^{K \times D}.
$
It does so by sampling a point cloud
$
\Pc = \{(p_i, n_i)\}_{i=1}^N
$
from the shape, where $p_i \in \xb$ is a surface point and $n_i \in \mathbb{S}^2$ its corresponding normal.
Then, it further subsamples $K \ll N$ points from $\Pc$ using Farthest Point Sampling (FPS), uses them to initialize the tokens,  and then uses cross-attention to pool information from all points $\Pc$ into the tokens.
The decoder $\Dc$, another transformer, evaluates the Signed Distance Function (SDF) $\Dc(q \mid \zb) \in \mathbb{R}$ of the 3D shape $\xb$ at any query location $q \in \mathbb{R}^3$ by attending to the latent tokens $\zb$.

Given this latent space, \vecset models the conditional distribution $p(\zb \mid I)$ over latent codes instead of the distribution $p(\xb \mid I)$ over 3D shapes.
Following~\cite{li24craftsman:,li25triposg:}, we do so by using a Diffusion Transformer (DiT)~\cite{peebles23scalable}, which learns a denoiser neural network $\zb_{t-1} = \Phi(\zb_t \mid I, t)$ that takes as input a noised version $\zb_t$ of the latent code $\zb$ and partially denoises it to $\zb_{t-1}$, conditioned on the image $I$ and step $t$.
The image $I$ is tokenized using a pre-trained image encoder (DINOv2~\cite{oquab2023dinov2}) and integrated into the DiT through cross-attention layers.

\subsubsection{From Generation to Regeneration.}%
\label{sec:injecting-the-3d-shape-condition}

We now generalize the model above to sample the distribution
$
p(\xb^H \mid \yb^L, I).
$
As noted in \cref{sec:intro}, while the meaning of the conditioning signal $\yb^L$ differs depending on the task, they can all be written as 3D shapes $\yb^L\subset \Rb^3$.
Hence, we propose to use the same \vecset representation for $\yb^L$ as well, and to learn the conditional distribution
$
p(\zb^H \mid \zb^L, I)
$
where $\zb^H$ and $\zb^L$ are the latent codes of $\xb^H$ and $\yb^L$, respectively.

Utilizing the same latent space for both the conditioning and generated signals $\zb^L$ and $\zb^H$ opens up a very simple yet effective design for the DiT 
$
\zb_{t-1}^H = \Phi(\zb_t^H \mid \zb^L, I, t);
$
namely, we can simply \emph{concatenate} the conditioning tokens $\zb^L$ to the noisy tokens $\zb^H_t$ and pass them together as input to the DiT\@.
In the experiments, we show that this is more effective than alternative designs where other encodings are used, including designs that use cross-attention to pass the conditioning tokens rather than concatenation.

\paragraph{Fewer tokens for coarser shapes.}

A useful property of \vecset, which we do not believe was noted or used before, is that the number of latent tokens $K$ can be adjusted according to the complexity of the encoded shape $\xb$.
In particular, because the conditioning shape $\yb^L$ contains little information, it can be encoded using far fewer tokens $\zb^L = \Ec(\Pc^L;C)$ than the generated shape $\xb^H$, where $C \ll K$.
Due to the quadratic complexity of self-attention, this significantly improves the model's efficiency.
Note that sparse voxel grids~\cite{xiang24structured}, also popular in 3D generation, do not have this property as a high voxel resolution is still required to encode accurately even a simple shape.

\paragraph{Token pre-processing.}

The authors of~\cite{chen25autopartgen} recently noted that concatenating two \vecset{}s approximates merging the two underlying 3D shapes, which differs from the regenerative effect we aim to achieve.
We thus pre-process the conditioning tokens $\zb^L$ before concatenation so that the model can better distinguish them from the noisy latent tokens $\zb^H$.
We do so by applying an MLP with zero-initialized weights, similar to ControlNet~\cite{zhang23adding}.
Zero-initialization avoids ``shocking'' the model when new control tokens $\zb^L$ are introduced.
We also find it useful to add positional embeddings to $\zb^L$ to help the model identify their spatial locations.
Formally, we can write our diffusion step as
$$
\zb_{t-1}^H = \Phi\Big(\zb_t^H \oplus \text{MLP}\big(\Ec(\Pc^L;C) + \text{PE}(\hat\Pc^L)\big) \Bigm| I, t\Big),
$$
where $\oplus$ denotes concatenation along the token dimension.
Note that only $\zb^H_t$ is denoised whereas $\zb^L$ remains fixed as a condition.

\begin{figure*}[t!]
\centering
\includegraphics[width=\linewidth]{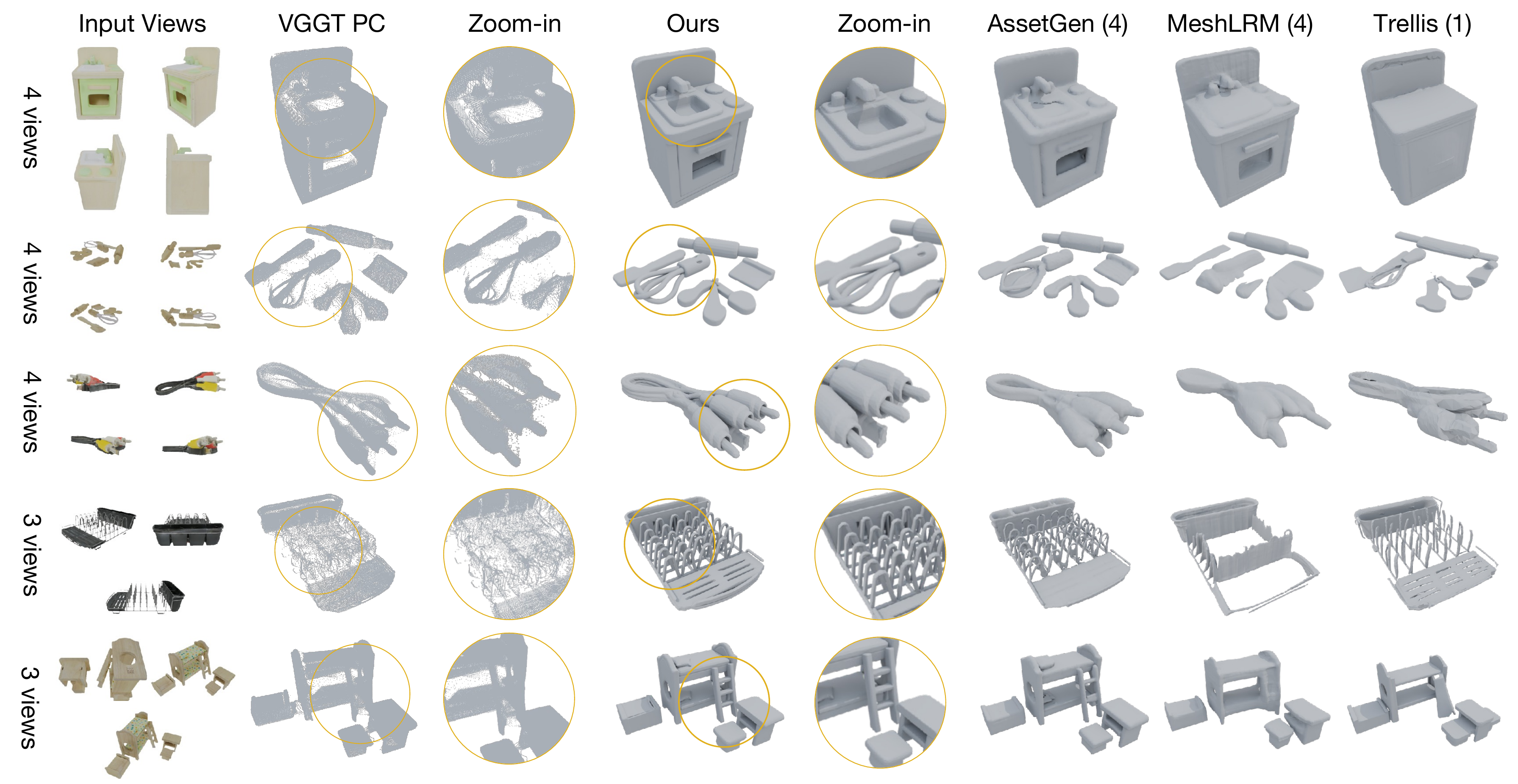}
\vspace{-0.6cm}
\caption{Qualitative comparisons for faithful Image-to-3D generation on the GSO dataset~\cite{downs22google}.
\name preserves the geometry of the initial VGGT point cloud~\cite{wang25vggt} while producing clean surfaces.
Baselines that rely solely on image conditioning tend to be less faithful and smooth out fine details.
For VGGT and \name, the number of input views is shown on the left; for the baselines, it is shown in parentheses.}%
\label{fig:vggt_gso_comparison}
\vspace{-0.3cm}
\end{figure*}

\subsubsection{Training Data.}%
\label{sec:training-objective}

In order to train the model to capture the conditional distribution
$
p(\zb^H \mid \zb^L, I)
$,
we require a dataset of triplets
$
\{(\zb_i^H, \zb_i^L, I_i)\}_{i=1}^n
$
comprising the encodings of high- and low-quality 3D shapes $\xb^H$ and $\xb^L$ along with the corresponding (optional) guidance images $I_i$.
The triplets must be specific to the different tasks one wishes to solve (e.g.,~enhancement, reconstruction, editing).
3D training data is already scarce, and procuring paired data for each task is much harder still.
Hence, we develop instead data engineering pipelines that construct such triplets automatically, starting from a collection of 3D objects $\xb^H$.
We detail this process for each target application in \cref{sec:applications}.

\begin{figure*}[t!]
\centering
\includegraphics[width=0.98\linewidth]{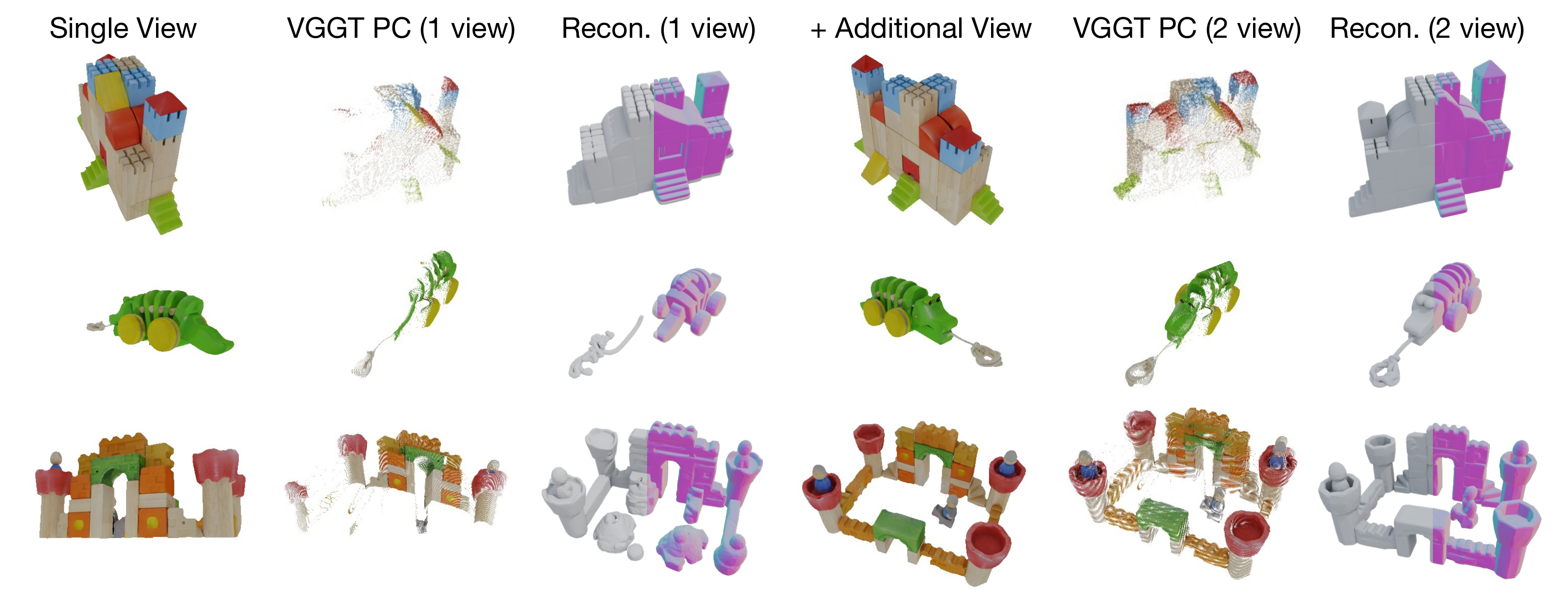}
\vspace{-0.3cm}
\caption{Reconstruction results of VGGT+\name from one and two images.
Note the improvement in the details on the side that is occluded in the first view.
The trend continues with more views, as shown by metrics in~\cref{table: gso}.}%
\label{fig:vggt_gso_12view}
\vspace{-0.5cm}
\end{figure*}

\subsection{Applications}%
\label{sec:applications}

We introduce several example applications of \name to show that the same model architecture and 3D dataset (\cref{sec:implementation}) can be used for multiple purposes.

\subsubsection{Base dataset.}%
\label{sec:base-dataset}

All our applications are trained from the \emph{same} base 3D dataset $\Dc$ as we only change how the data is automatically augmented in pre-processing.
The dataset comprises approximately 1M collection of artist-authored 3D objects very similar in spirit to Objaverse~\cite{deitke23objaverse:}.

\subsubsection{Compositional Shape Enhancement.}%
\label{sec:enhancement}

In \emph{shape enhancement}, we map a low-quality 3D shape $\yb^L$ to a high-quality one $\xb^H$.
We apply this idea to a \emph{compositional} setting~\cite{liu24part123:,chen25partgen,chen2025autopartgen,yang25holopart:,wang26worldgen:}, where a large 3D object is first decomposed and then upgraded one component at a time.
We are motivated by the fact that 3D generators operate at a finite resolution, and even high resolution ones like Sparc3D~\cite{li25sparc3d:} fail to generate sufficient details for large 3D objects such as whole scenes in one go.

\paragraph{Data Construction.}

Please see the Appendix for an illustrative overview.
To simulate the compositional (re){}generation process, we start by arranging random ground-truth (and thus high-quality) 3D objects $\xb^H \in \Dc$ in simple $k \times k$ grids.
These pseudo-compositions are then normalized (as a whole) in the $[-1,1]^3$ cube and encoded using \vecset into tokens.
Inspired by SDEdit~\cite{meng22sdedit:}, given a pre-trained DiT, we add noise to these tokens at level $t^*<T$ and denoise them back to $t=0$.
The 3D scene is then decoded from the denoised tokens and the degraded 3D shape $\yb^L$ of each object is extracted from the resulting geometry.
Increasing $k$ means that more objects are represented by the same number of tokens, controlling how much information is removed per-object, which also determines the final number of tokens $\zb^L$ we use to represent $\yb^L$.
$\zb^H$ is instead obtained by encoding the original object $\xb^H$ with the full token count. 

\paragraph{Generality.}

While we simulate the degradation process due to generating multiple objects together, in practice, as we show in the Appendix, our technique results in pairs $(\zb^H,\zb^L)$ that are representative of degradations common in other applications too, more so than, say, applying generic 3D smoothing filters~\cite{schroeder1992decimation, herrmann1976laplacian, sorkine2004laplacian} to $\xb^H$.
As a result, the resulting model can also be potentially applied to a wide range of degradations, including low-quality 3D assets (e.g., from old games) and models affected by lossy 3D compression.

\subsubsection{Guided Image-to-3D.}%
\label{sec:reconstruction}

In \emph{guided image-to-3D} we map an initial sparse but accurate 3D point cloud $\yb^L$ reconstructed from one or more images $\{I_1,\dots,I_L\}$ into a full 3D object $\xb^H$ faithful to the input data.
We are motivated by the fact that current image-to-3D generators produce high-quality and detailed 3D shapes that are however not necessarily faithful to the input image, and/or cannot account for more than one image as input.
On the other hand, large 3D reconstruction neural networks like VGGT~\cite{wang25vggt} and many others~\cite{wang25p3,keetha25mapanything:,lin25depth} reconstruct the visible geometry as faithfully as possible from several views, resulting in 3D reconstructions $\yb^L$ that are accurate, but are incomplete due to occlusion, lack details, and can be noisy.

\paragraph{Data Construction.}

Given a ground-truth 3D object $\xb^H$, we render 1--4 random views $\{I_1,\dots,I_L\}$ and pass them to VGGT to obtain a sparse reconstruction $\yb^L$.
For rendering, camera poses are sampled uniformly in azimuth, focal length, and with elevations in $[-10^\circ, 30^\circ]$.
In order to apply $\Ec$ to encode $\yb^L$, we also require the normals for each surface point, which VGGT does not provide, so we estimate them using covariance analysis~\cite{mitra2003estimating}.
Because the scale and pose of $\yb^L$ is arbitrary, we align it to $\xb^H$.
The conditioning image $I$ is randomly chosen from the rendered views and the triplet $(\zb^H, \zb^L, I)$ is added to the training dataset.

\subsubsection{3D Shape Editing.}%
\label{sec:editing}

In \emph{3D shape editing} we map an initial 3D shape $\yb^L$ into a modified shape $\xb^H$ that reproduces $\yb^L$ except for a specific 3D region $M$ which is edited to match a guide image $I$. The 3D masks can be as simple as a box or ellipsoid that fully contains the region of geometry we want to edit.
Following~\cite{gao20253d,li2025voxhammer}, the image $I$ is obtained automatically by (1) rendering a view of the original object $\yb^L$, (2) masking the 2D region corresponding to the 3D edit $M$, and (3) using a text-to-2D model~\cite{labs25flux.1} to inpaint the masked detail based on textual instructions.

\paragraph{Data Construction.}

To generate training data for this model we need to simulate the editing process.
The target 3D object $\xb^H$ is sampled as usual from the base dataset $\Dc$, the image $I$ is a render of this object, and $\zb^H$ is the encoding of $\xb^H$.
Then, it only remains to show how to construct the shape $\yb^L$ (and corresponding code $\zb^L$) that represents the \emph{source} of the edit.

To this end, given $\xb^H$, we first randomly sample an edit region $M$ as a 3D mask. The 3D masks can be as simple as a box or cylinder containing the geometry we want to edit.
Then we remove the original points within $M$ and the corresponding normals from $\Pc^H$ to form $\Pc^L$.
We obtain $\tilde{\zb} = \Ec(\Pc^L; C-k)$ and use $C-k < C$ tokens instead of all $C$ tokens.
The remaining $k$ edit tokens are used to tell the model where the edit should be applied and are initialized as learnable mask positional embeddings $\zb^M$ in latent space.
The final encoding for the degraded shape $\yb^L$ is the concatenation $\zb^L = \tilde{\zb} \oplus \zb^M$ of $\tilde{\zb}$ with $k$ learnable mask tokens $\zb^M$.
We thus obtain the final training triplet $(\zb^H, \zb^L, I)$.

The number of edit tokens $k$ represents the editing capacity.
It is adaptively set as $k =\max(C*r, k_{\min})$, where $r$ is the ratio of masked to total source points.
Capping $k$ with $k_{\min} \in [0,C)$ guarantees minimal editing capacity.
This adaptive allocation balances identity preservation with editing flexibility, enabling larger masks to produce greater geometric changes.

\vspace{-0.2cm}
\subsection{Implementation Details}%
\label{sec:implementation}
\vspace{-0.2cm}

Our base model is an in-house \vecset-based 3D generator, which is trained on the base dataset $\Dc$ and its multi-view renderings, in a similar manner to prior works ~\cite{li24craftsman:,li25triposg:,zhao25hunyuan3d,xiang24structured}.
This results in initial weights for both the VAE and the DiT\@.
We also apply the augmentation strategies described in \cref{sec:applications} to the base dataset to obtain the training datasets for each application.

The \textit{enhancement} DiT is trained by combining the augmentations for the enhancement models as well as the editing one.
Following the literature~\cite{tripo3d24text-to-3d, hunyuan3d25hunyuan3d, zhang20233dshape2vecset}, we train the DiT by progressively increasing the number of tokens $K = 512, 1024, 2048$ for 49K, 112K, and 390K steps.
To train the \textit{reconstruction} DiT, we fine-tune the enhancement model on the VGGT outputs for 25K steps.
To train the \textit{Shape editing} DiT we used the masked training set and $K{=}2048$ high-quality tokens.
Training converges in just 25K steps.
For every model, we used $C=512$ \emph{low-quality} conditioning tokens.

For additional robustness, we further augment the conditioning 3D shape by adding Gaussian blobs imitating floaters and by perturbing point normals, and we augment the image conditioning via random crops, color jitter, random backgrounds, and blur.
We adopt the $v$-prediction~\cite{salimans22progressive} with a scaled linear schedule and $p=0.1$ image-condition dropout.
For inference, we use DDPM sampling with 100 steps and a CFG scale of 5.
Please see the Appendix for more details.

\section{Results}%
\label{sec:results}
\vspace{-0.4cm}

We evaluate the effectiveness of \name on the applications of \cref{sec:applications}.

\subsection{Compositional 3D Enhancement}%
\label{sec:results_enhancement}

\begin{table}[t!]
\centering
\caption{3D shape enhancement evaluation.
MV-ImageReward~\cite{xu2024imagereward} reports the average ImageReward over four renders per object.}%
\label{table: refine}
\vspace{-0.3cm}
\resizebox{0.7\linewidth}{!}{
\begin{tabular}{c c c c}
\toprule
{Metrics} & {Input~\cite{li25sparc3d:}} & {$+$DetailGen3D~\cite{deng2024detailgen3d}} & {$+$Ours} \\
\midrule
{ULIP-3D~\cite{xue2023ulip}} & {0.2280} & {0.2294} & {\textbf{0.2626}} \\
\midrule
{MV-ImageReward~\cite{xu2023imagereward}} & {0.1716} & {0.1003} & {\textbf{0.3394}} \\
\bottomrule
\end{tabular}
}
\vspace{-0.4cm}
\end{table}

We first evaluate \name on the task of compositional 3D enhancement.
We consider a practical scenario where the goal is to generate complex 3D scenes.
Here, the starting point is an image $I$ (concept art) of the entire scene, which is then reconstructed as a 3D object by an image-to-3D model \cite{li25sparc3d:}.
This results in a 3D scene that is globally consistent but lacks detail and fidelity.
We use an off-the-shelf 3D part segmenter~\cite{chen2025autopartgen} to decompose the scene into individual objects $\yb^L$ and then use \name to upgrade them to high-quality versions $\xb^H$.
To do so, we also require a guide image $I'$ that captures the details of each object that needs upgrading.
We obtain the latter automatically using an off-the-shelf VLM, as explained in the Appendix.

Qualitative results are given in \cref{fig:scene_refine}.
Quantitatively, we evaluate the performance of the model on 623 decomposed low-quality objects extracted from 21 scene meshes.
We use ULIP-3D~\cite{xue2023ulip} to measure 3D shape-view alignment and Multi-View (MV) ImageReward~\cite{xu2023imagereward} to measure perceptual quality.
Specifically, we generate concise text captions of conditional images with LLaMA\,3.2~\cite{grattafiori2024llama} and use these prompts to compute the averaged ImageReward across four rendered views per object.
As shown in \cref{table: refine} and \cref{fig:obj_refine}, \name substantially improves geometric fidelity and visual quality of initial decomposed objects~\cite{li25sparc3d:}, surpassing state-of-the-art baselines~\cite{deng2024detailgen3d} for geometry refinement.

\begin{table}[t!]
\caption{3D object reconstruction on GSO~\cite{downs22google}.
``\# View'' denotes the number of input images per baseline.
VGGT results are based on meshes converted by Poisson reconstruction~\cite{kazhdan2006poisson}.}%
\label{table: gso}
\vspace{-0.3cm}
\centering
\resizebox{\linewidth}{!}{%
\begin{tabular}{@{}ccccccccc@{}}
\toprule
& Model & \# View & CD ($\downarrow$) & F-score ($\uparrow$) & IoU ($\uparrow$) & PSNR ($\uparrow$) & SSIM ($\uparrow$) & LPIPS ($\downarrow$) \\ \midrule
\multirow{2}{*}{} & InstantMesh~\cite{xu2024instantmesh} & 1 & 0.0188 & 0.2840  & 0.6025  & 20.3317 & 0.9188 & 0.1389 \\ \cmidrule(l){2-9}
& TRELLIS~\cite{xiang2025structured}   & 1 & 0.0122 & 0.4216 & 0.6442 & 22.4605 & 0.9342  & 0.1059 \\ \midrule
\multirow{3}{*}{\rotatebox[origin=c]{90}{\shortstack{View\\$>$\,1}}}  & AssetGen~\cite{siddiqui2024meta} & 4 & 0.0111  & 0.4663 & 0.6703 & \textbf{24.3809} & 0.9206 & 0.1024    \\ \cmidrule(l){2-9}
& TRELLIS~\cite{xiang2025structured} & 4 & 0.0123 & 0.4318 & 0.6552  & 22.7584  & 0.9351  & 0.1036 \\ \cmidrule(l){2-9}
& MeshLRM~\cite{wei2024meshlrm} & 4 & 0.0115  & 0.4237 & 0.7035 & 22.8111 & 0.9290 & 0.1180 \\ \midrule
\multirow{8}{*}{\rotatebox[origin=c]{90}{Hybrid}}   & VGGT~\cite{wang2025vggt} & 1 & 0.1286  & 0.0087 & 0.0247 & 9.0717  & 0.7979  & 0.5356  \\ \cmidrule(l){2-9}
& \cellcolor{our_green} + \name  & \cellcolor{our_green} 1  & \cellcolor{our_green} {0.0135}  & \cellcolor{our_green} {0.3658}  & \cellcolor{our_green} {0.6438}  & \cellcolor{our_green} {21.8982}  & \cellcolor{our_green} {0.9315}  & \cellcolor{our_green} {0.1097} \\ \cmidrule(l){2-9}
& VGGT~\cite{wang2025vggt}  & 2 & 0.1340 & 0.0083 & 0.0278 & 8.8537  & 0.7867  & 0.5695  \\ \cmidrule(l){2-9}
& \cellcolor{our_green} {+ \name}  & \cellcolor{our_green} 2  & \cellcolor{our_green} {0.0112}  & \cellcolor{our_green} {0.4299} & \cellcolor{our_green} {0.7053} & \cellcolor{our_green} {22.9795} & \cellcolor{our_green} {0.9353}  & \cellcolor{our_green} {0.0985}  \\ \cmidrule(l){2-9}
& VGGT~\cite{wang2025vggt} & 3 & 0.1435 & 0.0089  & 0.0274 & 8.3342 & 0.7615  & 0.6077  \\ \cmidrule(l){2-9}
& \cellcolor{our_green} {+ \name}  & \cellcolor{our_green} 3 & \cellcolor{our_green} {0.0090}  & \cellcolor{our_green} {0.4650} & \cellcolor{our_green} {0.7372} & \cellcolor{our_green} {23.6394}  & \cellcolor{our_green} {0.9385}  & \cellcolor{our_green} {0.0918} \\ \cmidrule(l){2-9}
& VGGT~\cite{wang2025vggt} & 4 & 0.1276  & 0.0119 & 0.0193 & 8.7823 & 0.7704 & 0.5952 \\ \cmidrule(l){2-9}
& \cellcolor{our_green} {+ \name}  & \cellcolor{our_green} 4 & \cellcolor{our_green} \textbf{0.0081} & \cellcolor{our_green} \textbf{0.4913} & \cellcolor{our_green} \textbf{0.7574}  & \cellcolor{our_green} \underline{24.2754}  & \cellcolor{our_green} \textbf{0.9408}  & \cellcolor{our_green} \textbf{0.0873} \\
\bottomrule
\end{tabular}%
}
\vspace{-0.5cm}
\end{table}

\subsection{Guided Image-to-3D}%
\label{sec:results_reconstruction}

We consider 3D generation guided by the output of a reconstruction network, VGGT\@.
Following~\cite{siddiqui24meta, wei2024meshlrm}, we benchmark the reconstruction performance on the GSO dataset~\cite{downs22google}.
For perceptual evaluation, we compute PSNR, SSIM~\cite{wang04image}, and LPIPS~\cite{zhang2018unreasonable} between the rendered and ground-truth normal maps by rescaling unit normals to $[0, 255]$.
For geometric accuracy, we report Chamfer Distance (CD), F-score@1$\%$, and IoU.

We compare against both single-view baselines (InstantMesh~\cite{xu2024instantmesh}, TRELLIS~\cite{xiang2025structured}) and multi-view methods (3D AssetGen~\cite{siddiqui24meta}, TRELLIS (multi-image variant)~\cite{xiang2025structured}, and MeshLRM~\cite{wei2024meshlrm}).
TRELLIS (multi-image) randomly selects one of the conditioning images at each sampling step.
We also evaluate the unrefined VGGT reconstructions $\xb^L$ as a separate series of baselines, for which we convert them to meshes using Poisson reconstruction~\cite{kazhdan2006poisson}.

As shown in \cref{table: gso,fig:vggt_gso_comparison}, \name significantly improves the geometric accuracy over the baselines.
Single-view methods suffer from inaccurate hallucinations due to limited visibility.
Moreover, \name matches or even outperforms specialized multi-view methods by using the same amount or even fewer input views (i.e., two to three views).
\cref{fig:vggt_gso_12view} demonstrates robust reconstruction from sparse views.
With a single image, it reconstructs fine surface details of visible regions and progressively refines occluded areas as more views are provided, highlighting the synergy between feed-forward prediction and generative restoration.

\begin{table}[t!]
\caption{Ablation study of the conditioning mechanism.
Designs of $\text{ID}=1$ and $\text{ID}=2$ are inspired by CLAY~\cite{zhang24clay:} and Hunyuan3D-Omni~\cite{hunyuan3d2025hunyuan3d}, respectively.}%
\label{table: ablation_condition}
\vspace{-0.3cm}
\centering
\resizebox{\linewidth}{!}{
\begin{tabular}{c c c c c c c c c}
\toprule
{Id} & {Method} & {Cond. Space} & CD ($\downarrow$) & F-score ($\uparrow$) & IoU ($\uparrow$) & PSNR ($\uparrow$) & SSIM ($\uparrow$) & LPIPS ($\downarrow$) \\
\midrule
1 & {Additional CA} & {CA} & {0.0066} & {0.6107} & {0.6889} & {24.4104} & {0.9258} & {0.0900} \\
\midrule
2 & {Extended CA} & {CA} & {0.0271} & {0.2374} & {0.2818} & {18.7145} & {0.9051} & {0.1410} \\
\midrule
3 & {Add $\zb^L$} & {Input} & {0.0112} & {0.5259} & {0.5960} & {22.7384} & {0.9175} & {0.1060} \\
\midrule
4 & {Ours} & {Input} & {\textbf{0.0044}} & {\textbf{0.7538}} & {\textbf{0.7735}} & {\textbf{26.3958}} & {\textbf{0.9338}} & {\textbf{0.0765}} \\
\bottomrule
\end{tabular}
}
\vspace{-0.5cm}
\end{table}

\begin{figure*}[t!]
\centering
\includegraphics[width=\linewidth]{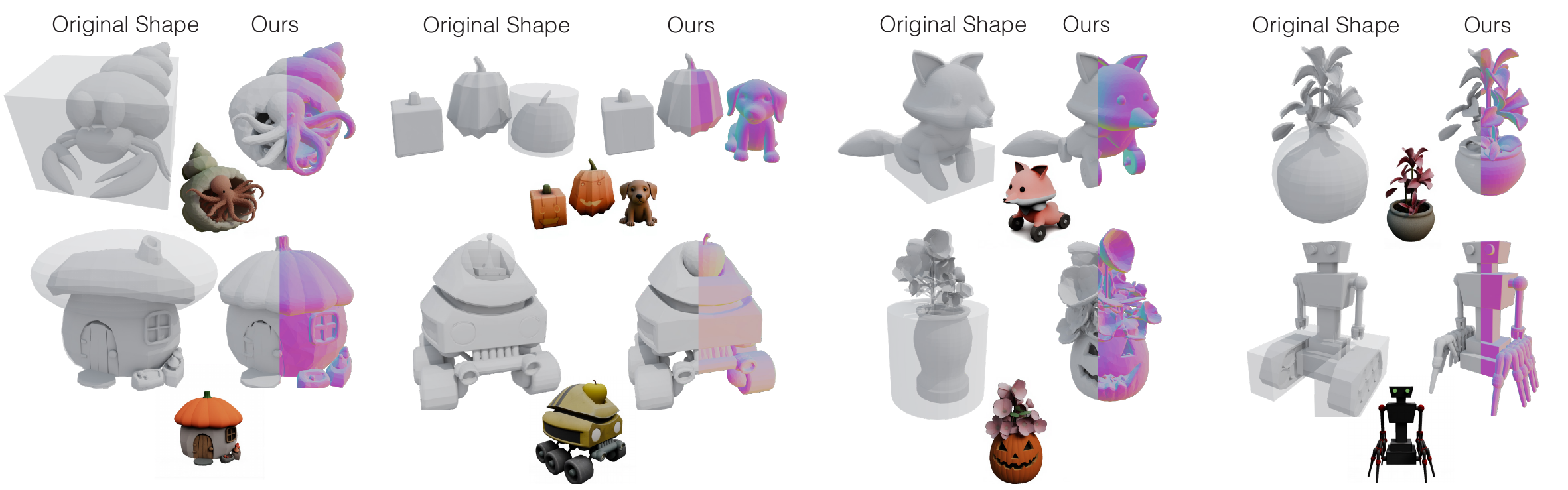}
\vspace{-0.6cm}
\caption{3D object editing examples.
Original shapes and edited images (shown in inset) are from~\cite{li2025voxhammer}.
The resulting shape maintains both the orientation and the level of detail of the input shape. The 3D edit $M$ masking the edit region is also visualized.}%
\label{fig:obj_edit}
\vspace{-0.4cm}
\end{figure*}

\begin{figure}[t!]
    \centering
    \includegraphics[width=\linewidth]{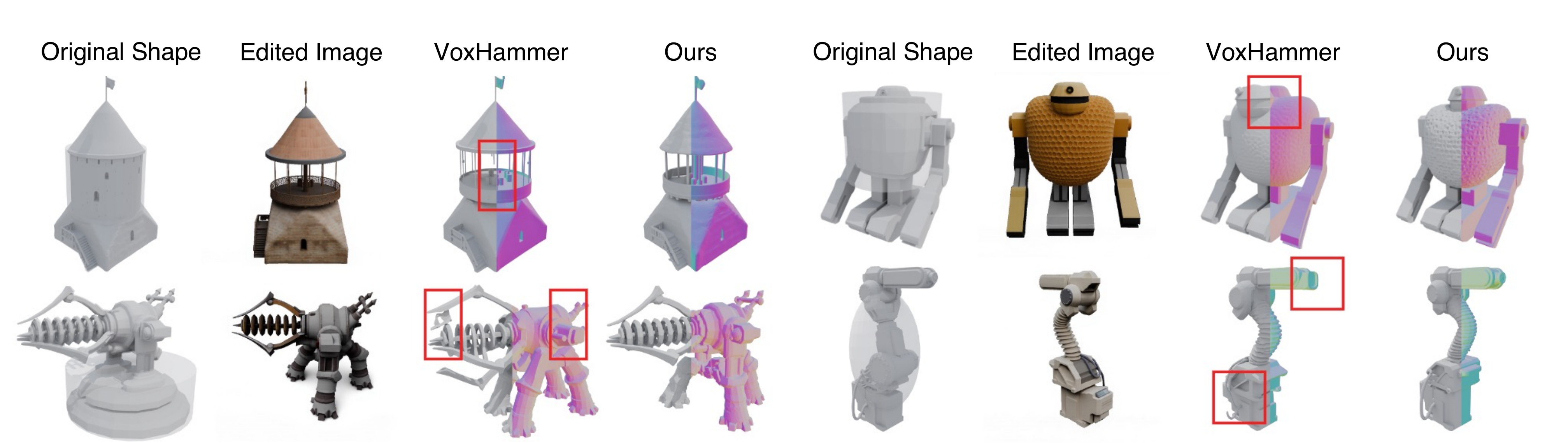}
    \vspace{-0.4cm}
    \caption{Qualitative comparisons of object editing with~\cite{li2025voxhammer}.
    Red boxes highlight either artifacts present in the editing results or differences between the edited shape and the original shape. The 3D edit $M$ masking the edit region is also visualized.}%
    \label{fig:obj_edit_comparison}
    \vspace{-0.4cm}
\end{figure}
\begin{figure}[t]
    \centering
    \includegraphics[width=0.95\linewidth]{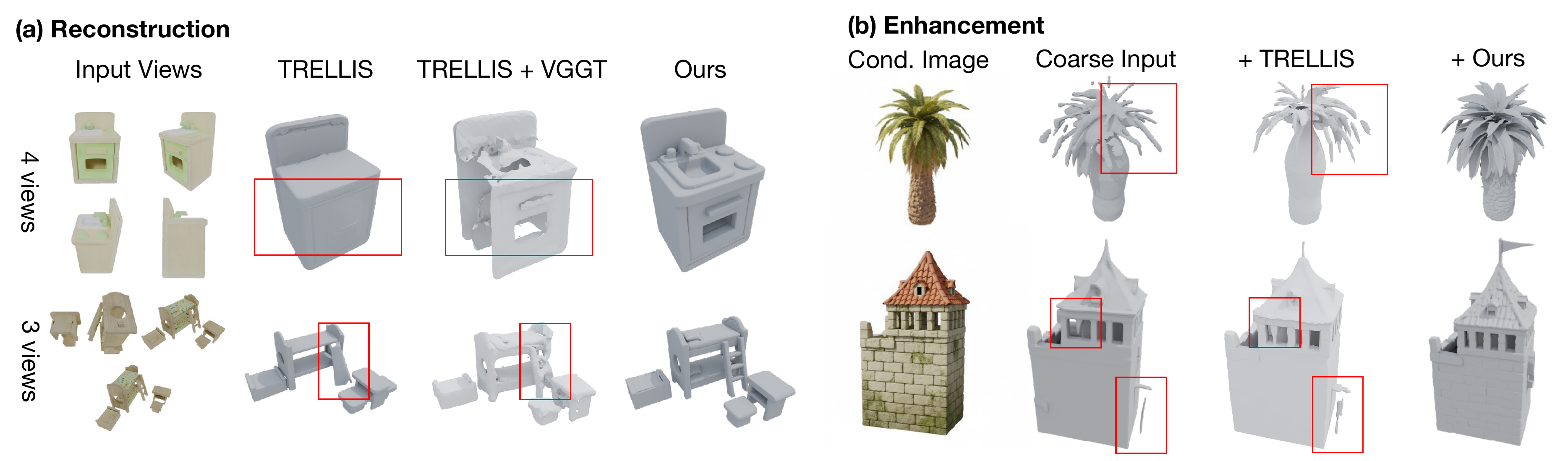}
    \vspace{-0.4cm}
    \caption{TRELLIS results on regeneration tasks. (\textbf{a}) Reconstruction from sparse views. We treat VGGT predictions as sparse structures in TRELLIS’s first stage (TRELLIS + VGGT). (\textbf{b}) Enhancement from degraded geometry. TRELLIS struggles to correct artifacts. Both TRELLIS and ours use exactly the same input shape or VGGT predictions.}
    \label{fig:trellis}
    \vspace{-0.4cm}
\end{figure}
\begin{figure}[t!]
\centering
\includegraphics[width=0.9\linewidth]{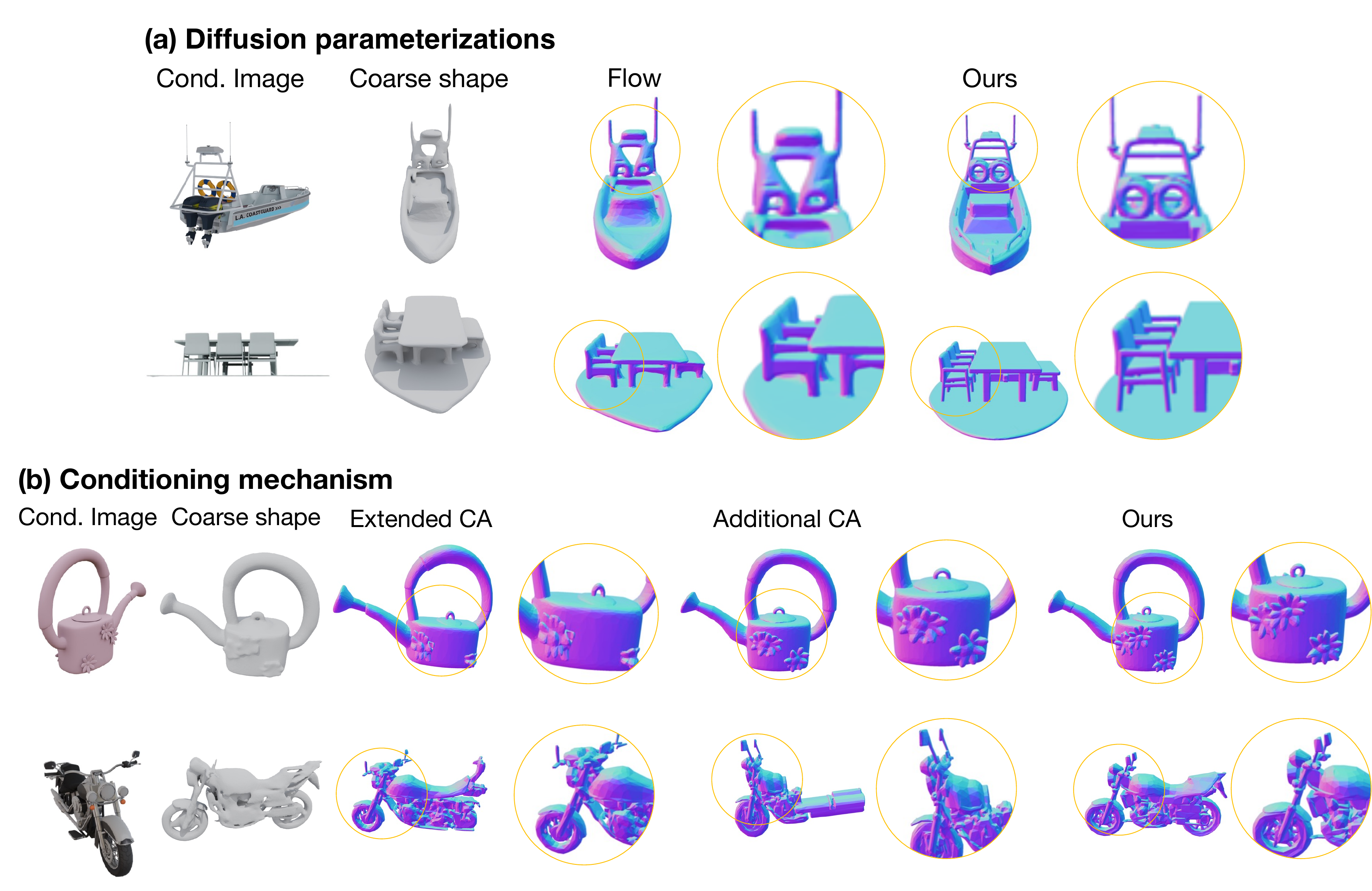}
\vspace{-0.4cm}
\caption{Qualitative examples from the ablation study.
(\textbf{a}) Samples from the ablation study of different diffusion parameterizations in~\cref{table: ablation_diffusion}.
(\textbf{b}) Samples from the ablation study of the conditioning mechanism in~\cref{table: ablation_condition}.
Designs of \textit{Extended CA} and \textit{Additional CA} are inspired by~\cite{hunyuan3d2025hunyuan3d} and~\cite{zhang24clay:}, respectively.
The proposed method consistently outperforms other design choices with fine-grained details.}%
\label{fig:ablation}
\vspace{-0.6cm}
\end{figure}

\vspace{-0.3cm}
\subsection{3D Shape Editing}

To assess 3D shape editing, we utilize 3D source objects, editing masks, and image prompts from the Edit3D-Bench dataset~\cite{li2025voxhammer}.
The image prompts were obtained by~\cite{li2025voxhammer} by rendering the masked source view and inpainting them using FLUX.1 Fill~\cite{labs25flux.1} conditioned on the intended target text prompt.

As shown in \cref{fig:obj_edit}, our framework enables effective 3D shape editing without architectural modifications.
\cref{fig:obj_edit_comparison} shows that \name can capture better fine-grained and large-scale structural changes compared to~\cite{li2025voxhammer}, while preserving the original source identity.
Unlike prior methods requiring expensive inversion~\cite{li2025voxhammer} or post-optimization~\cite{liu2024make, sella2023vox, zhuang2024tip}, our approach is computationally efficient and scalable.
Moreover, by operating directly in the 3D \vecset space, it inherently preserves multi-view coherence without the need for 2D-to-3D lifting~\cite{chen2024generic, barda2025instant3dit}.

\begin{table}[t!]
\caption{Ablation study of different diffusion parameterizations.
The design of ``Flow'' is inspired by~\cite{deng24detailgen3d:}.}%
\label{table: ablation_diffusion}
\vspace{-0.4cm}
\centering
\resizebox{0.85\linewidth}{!}{
\begin{tabular}{c c c c c c c}
\toprule
{Method} & CD ($\downarrow$) & F-score ($\uparrow$) & IoU ($\uparrow$) & PSNR ($\uparrow$) & SSIM ($\uparrow$) & LPIPS ($\downarrow$) \\
\midrule
{Flow} & {0.0061} & {0.6026} & {0.6594} & {24.6295} & {0.9285} & {0.0893} \\
\midrule
{Ours} & {\textbf{0.0051}} & {\textbf{0.7003}} & {\textbf{0.6961}} & {\textbf{25.4428}} & {\textbf{0.9298}} & {\textbf{0.0787}} \\
\bottomrule
\end{tabular}
}
\vspace{-0.7cm}
\end{table}

\subsection{Ablation Studies}
\vspace{-0.2cm}

\paragraph{Conditioning Mechanism.}
A key design choice (\cref{sec:method}) is the mechanism for conditioning the generator on the low-information shape $\yb^L$.
We compare \name's approach to alternative strategies:
(1)~introducing a separate cross-attention block dedicated to the 3D condition, parallel to the image-conditioned attention block;
(2)~extending the existing image-conditioned attention by concatenating coarse point cloud embeddings with DINO image features; and
(3)~directly adding $\zb^L$ to the noisy latents.
Designs (1) and (2) are inspired by CLAY~\cite{zhang24clay:} and Hunyuan3D-Omni~\cite{hunyuan3d2025hunyuan3d}, respectively.
In \cref{table: ablation_condition,fig:ablation}, our design outperforms baselines, potentially due to its flexible learning capacity.

\paragraph{Diffusion Parameterization.}
We further compare our diffusion formulation with a rectified flow baseline~\cite{lipman2023flow, liu2023flow}, where the model directly maps coarse latents $\zb^L$ to the ground-truth latent distribution without explicit Gaussian noise. This alternative choice is inspired by DetailGen3D~\cite{deng2024detailgen3d}.
\cref{table: ablation_diffusion} shows that our diffusion parameterization achieves better reconstruction on the validation set.

\paragraph{Applicability of existing generation frameworks to \textbf{re}generation.} 
We further test whether existing (controllable) 3D \textit{generation} frameworks are readily generalizable to \textit{regeneration} tasks. First, we apply TRELLIS to enhancement and reconstruction by respectively encoding degraded geometry or VGGT predictions into sparse structures in its first stage, followed by SLAT generation in the second stage. As shown in \cref{fig:trellis}, TRELLIS fails to recover fine geometry, as it is trained on \textit{clean geometry} which does not generalize to broken, incomplete inputs. Similarly, although CLAY \cite{zhang24clay:} is not publicly available, we evaluate Hunyuan3D-Omni \cite{hunyuan3d2025hunyuan3d}, a closely related \textit{concurrent} framework, on reconstruction tasks and observe similar performance degradation (see Appendix). These results suggest that existing \emph{generation} frameworks do not generalize to regeneration tasks with degraded/incomplete geometry without fine-tuning, motivating dedicated regeneration formulations and data.

Overall, using official baselines (when available) and controlled reimplementations under a fixed backbone and dataset, we show that our improvements stem from both dataset design \textit{and} a generalizable architecture.

\vspace{-0.5cm}
\section{Conclusion}
\vspace{-0.3cm}

We have introduced \name, a versatile 3D regenerator that unifies diverse 3D-to-3D tasks under a single framework.
Future work includes extending the framework to dynamics~\cite{ShapeGen4D} and incorporating richer control signals, such as skeletons for rigged assets. Regarding limitations, while \name demonstrates strong controllability, exploring alternative 3D latent representations (e.g., SLAT on active voxels~\cite{xiang2025structured}) may further improve generalizability.

{
    \small
    \bibliographystyle{splncs04}
    \bibliography{general,specific,main}
}

\clearpage
\appendix
\setcounter{page}{1}

\begin{center}
    {\Large \textbf{Supplementary Material of \name: A Unified 3D Geometry Regeneration Framework}}\\[1em]
\end{center}

The supplementary sections are organized as follows:
\begin{itemize}
    \item \Cref{supp sec: training_details} provides more experimental details, including architecture and training hyperparameters.
    \item \Cref{supp sec: enhancement} analyzes the data construction pipeline of 3D shape enhancement task.
    \item \Cref{supp sec: object image generation} provides additional details on our method for refining the segmented per-object images, which serve as input for 3D enhancement in the context of object-based scene generation.
    \item \Cref{supp sec: qualitative results} presents additional qualitative results and comparison for each task.
    \item \Cref{supp sec: related works} expands the review of related work.
\end{itemize}
Other parts of the supplementary material include additional qualitative results of refining 3D assets, provided as a dynamic web page.
Please refer to the README file in the supplementary material for simple instructions on how to view it on a local machine.

\vspace{-0.2cm}
\section{Training Details}%
\label{supp sec: training_details}

\paragraph{Architecture.}
As described in the main paper, \name is built on a frozen pre-trained 3D VAE whose backbone is a 24-layer Transformer.
This VAE is trained by randomly sampling the latent token count $C \in \{512,1024,2048\}$.
For the DiT denoiser, the 3D condition is provided as \vecset latents, augmented with a zero-initialized learnable positional embedding.
The full model contains approximately 2.8B parameters.
Each Transformer block uses a hidden dimension of 2048 with 16 attention heads, 24 layers, 11 skip connections, and an image embedding dimension of 1536.
The input point cloud resolution is fixed to $N = 32{,}768$ points for both high-quality and low-quality shapes.

\paragraph{Training.}
We use the AdamW~\cite{loshchilov19decoupled} optimizer with a learning rate of $10^{-4}$ for $C = 512$ and $2\times 10^{-5}$ for $C \in \{1024, 2048\}$, with a warm-up from $10^{-6}$ during the first epoch.
Training follows the scaled linear time schedule of Stable Diffusion~\cite{rombach2022high, lin23common} with $\beta_{\text{start}} = 0.00085$ and $\beta_{\text{end}} = 0.0015$.

\begin{figure*}[t!]
\centering
\includegraphics[width=\linewidth]{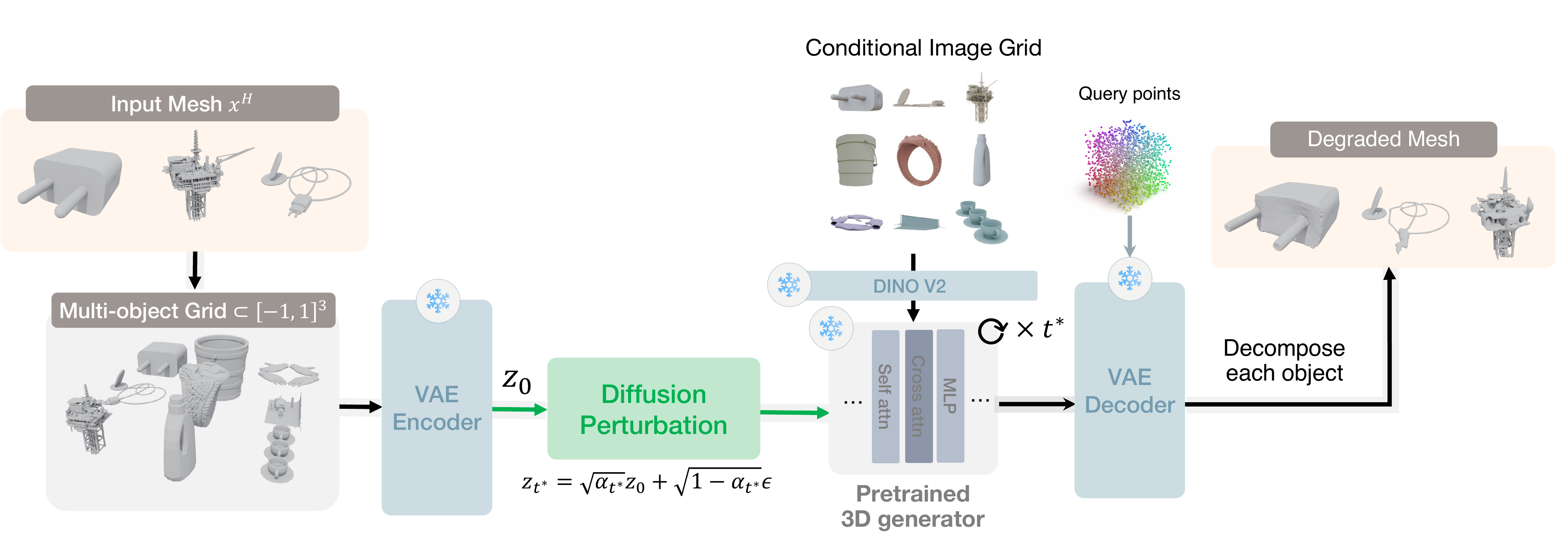}
\vspace{-0.3cm}
\caption{Data construction pipeline for 3D shape enhancement.
We synthesize compositional scenes by placing $\xb^H$ on a multi-object grid ($3\times3$ shown).
The scene is encoded into \vecset tokens $\z_0$, noised at $t^*=350$ (under a variance-preserving (VP) schedule~\cite{ho2020denoising, song21score-based}), and denoised to $t=0$ using a pre-trained 3D generator and the render of the grid as image conditioning.
Degraded shapes can then be extracted as clusters of the reconstructed shape.}%
\label{supp fig: data_pipeline}
\vspace{-0.2cm}
\end{figure*}
\begin{figure}[t!]
\centering
\vspace{-0.3cm}
\includegraphics[width=\linewidth]{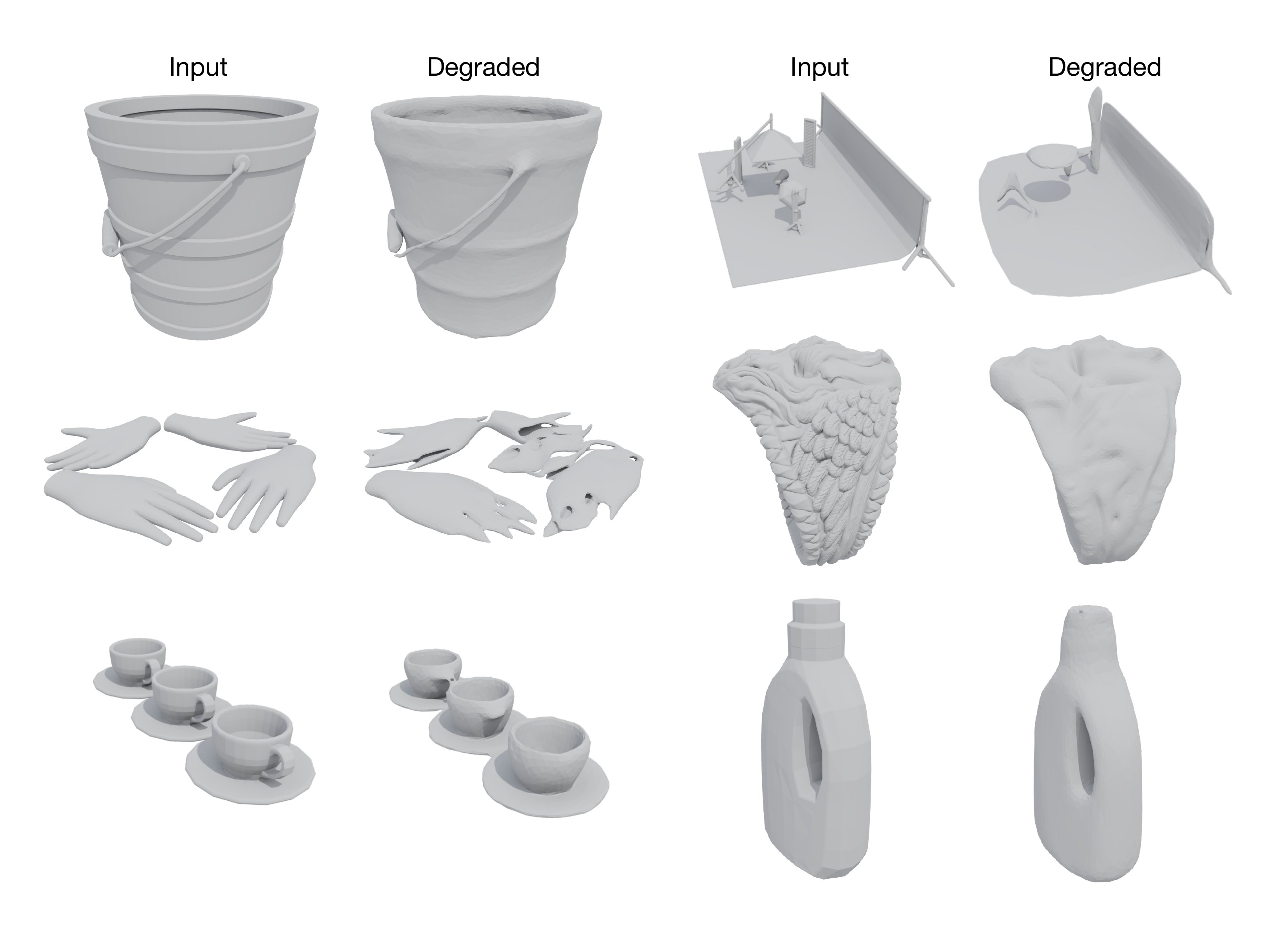}
\caption{Paired examples of ground-truth meshes $\xb^H$ and their degraded counterparts.
Each $\xb^H$ is randomly placed in a predefined $3 \times 3$ grid within the unit cube $[-1,1]^3$.}%
\label{supp fig: data_example}
\vspace{-0.3cm}
\end{figure}

\section{3D Shape Enhancement}%
\label{supp sec: enhancement}

We present the details of the overall dataset construction, including the production of the conditioning images. 

\paragraph{Dataset construction.}

\Cref{supp fig: data_pipeline} illustrates our data construction pipeline for 3D shape enhancement.
To simulate coarse objects arising in compositional 3D scene generation, we first construct synthetic multi-object scenes by placing high-quality meshes $\xb^H$ into predefined spatial slots.
The resulting scene is encoded into \vecset tokens.
Following the sampling process of compositional 3D generators, we perturb these tokens at a moderate noise level $t^* = 350 < T$ inspired by SDEdit~\cite{meng22sdedit:}.
The noisy scene tokens are then denoised using a corresponding image grid, where each object view is randomly selected with varying azimuth, elevation, and focal length.

We observe that this moderate corruption preserves the pose of the original objects in the degraded meshes.
The diffusion forward process first removes the high-frequency details while retaining a coarse geometric structure, which makes it well suited for paired data construction.

\begin{wrapfigure}{r}{0.5\textwidth}
\centering
\includegraphics[width=\linewidth]{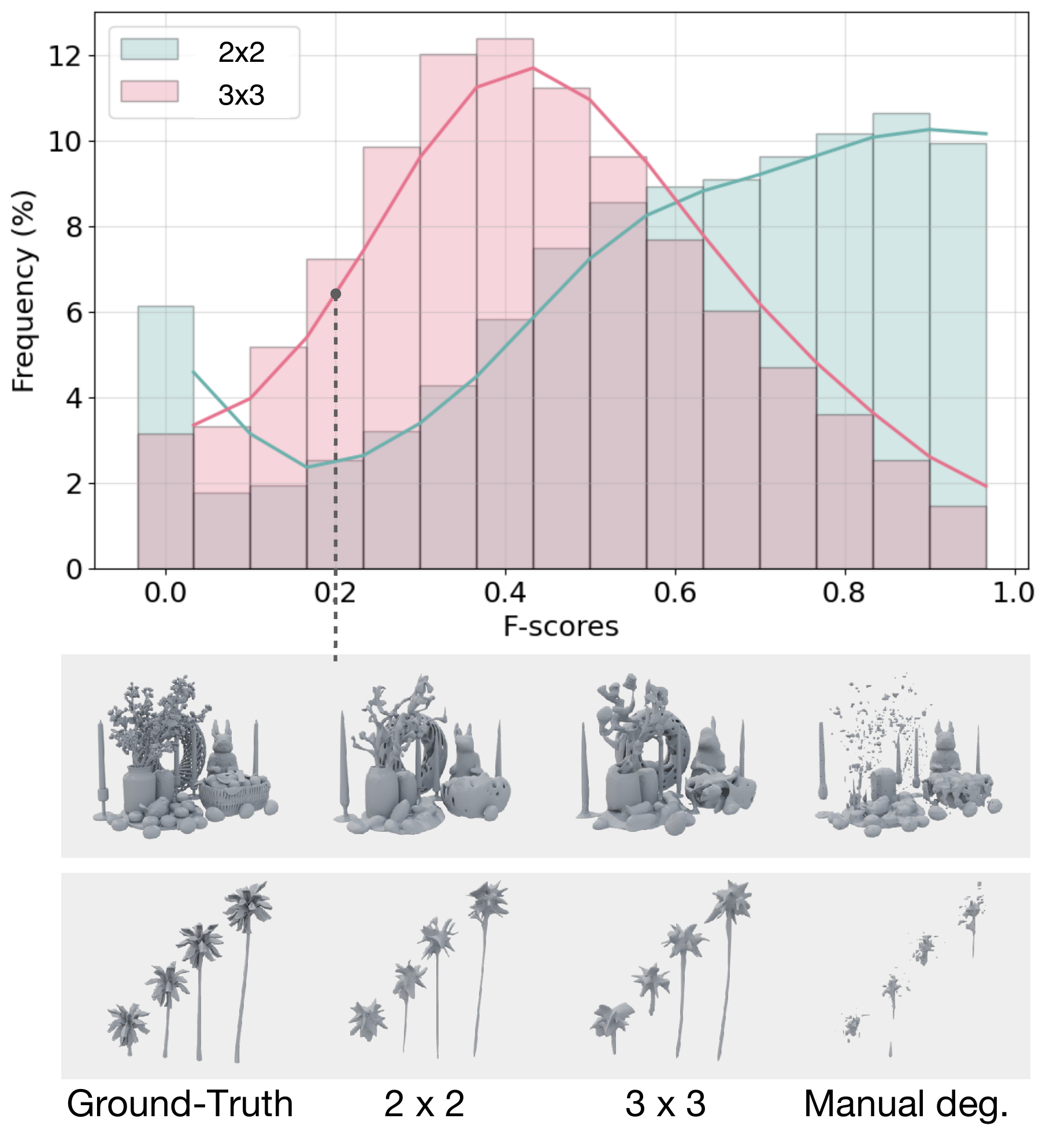}
\vspace{-0.8cm}
\caption{Histogram of F-score$@1\%$ over ground-truth and degraded meshes.
Simple manual degradations, such as low-pass filtering of the SDF or Taubin smoothing~\cite{taubin1995curve}, do not allow controlling the degradation consistently across objects with different geometric complexity, as similarly reported in~\cite{deng2024detailgen3d}.}%
\label{supp fig: fscore}
\vspace{-0.5cm}
\end{wrapfigure}

\paragraph{Analysis of degraded datasets.}

Using F-score$@1\%$, \cref{supp fig: fscore} further quantifies the severity of degradation as a function of grid size.
Increasing the grid resolution from $2\times2$ to $3\times3$ yields progressively stronger degradation due to the necessity to share the model capacity.
Thus, we adopt a progressive training strategy: degraded meshes with $2\times 2$ grids are used for the first stage ($C=512$) and $3\times 3$ grids for the later stages ($C \in \{1024,2048\}$).

\Cref{supp fig: data_example} shows that our pipeline produces more realistic and diverse degradation patterns compared to classical geometric degradations (e.g., Taubin smoothing or mesh simplification~\cite{taubin1995curve, schroeder1992decimation, herrmann1976laplacian, sorkine2004laplacian}).
Traditional operators often induce negligible changes in simple geometries or collapse structure in complex ones, as similarly reported in~\cite{deng24detailgen3d:}.
As another alternative, DetailGen3D~\cite{deng24detailgen3d:} adopts LRM reconstructions as low-quality proxies; using that strategy would result in the domain gap with respect to compositional 3D generation and lacks a principled way to control the severity of degradation.

\paragraph{Conditioning Mechanism.}
A key design choice (\cref{sec:method}) is how we condition the generator on the low-information shape $\yb^L$.
In \cref{table: ablation_condition} of the main paper, we compare \name against three alternative conditioning strategies: (1) introducing a separate cross-attention block dedicated to the 3D condition, parallel to the image-conditioned attention block; (2) extending the existing image-conditioned attention by concatenating coarse point-cloud embeddings with DINO image features; and (3) directly adding $\zb^L$ to the noisy latents.
Here, we run an additional ablation in which we condition \name on point clouds sampled from the coarse shape, rather than on coarse point-cloud embeddings. This setting is closest to Design (1), but adds extra point-cloud conditioning via the cross-attention block instead of using coarse point-cloud embeddings.

As shown in \cref{table: ablation_condition_extra}, our design still outperforms these baselines, highlighting the expressiveness of the point-cloud embeddings as conditioning signal.
\begin{table}[t!]
\caption{Ablation study of the conditioning mechanism.
Designs of $\text{ID}=1$ and $\text{ID}=2$ are inspired by CLAY~\cite{zhang24clay:} and Hunyuan3D-Omni~\cite{hunyuan3d2025hunyuan3d}, respectively.}%
\label{table: ablation_condition_extra}
\vspace{-0.3cm}
\centering
\resizebox{\linewidth}{!}{
\begin{tabular}{c c c c c c c c c}
\toprule
{Id} & {Method} & {Cond. Space} & CD ($\downarrow$) & F-score ($\uparrow$) & IoU ($\uparrow$) & PSNR ($\uparrow$) & SSIM ($\uparrow$) & LPIPS ($\downarrow$) \\
\midrule
1 & {Additional CA} & {CA} & {0.0066} & {0.6107} & {0.6889} & {24.4104} & {0.9258} & {0.0900} \\
\midrule
2 & {Additional CA with point cloud} & {CA} & { 0.0162} & {0.6051} & {0.4540} & {21.06} & {0.9126} & {0.1178} \\
\midrule
3 & {Ours} & {Input} & {\textbf{0.0044}} & {\textbf{0.7538}} & {\textbf{0.7735}} & {\textbf{26.3958}} & {\textbf{0.9338}} & {\textbf{0.0765}} \\
\bottomrule
\end{tabular}
}
\vspace{-0.5cm}
\end{table}

\paragraph{Remark on the necessity of an enhancement model.} 

In 3D enhancement, preserving consistency with the original measurements is crucial since any mismatch in rotation or scale between the coarse input and the enhanced output can break downstream applications. For example, if a model deviates from the original 6-DoF pose of the coarse part, it may not fit properly into the original composite object.
Single-image 3D generative models often fail to maintain this pose consistency because they lack explicit 3D conditioning or any other form of shape constraints.
Moreover, the ambiguous nature of generation conditioned on a single view means that these models may miss important detail or hallucinate structures that do not match the true geometry, especially in occluded regions. Relevant examples are provided in the supplementary web page.

\section{Per-Object image generation for 3D Shape Enhancement}%
\label{supp sec: object image generation}
\begin{figure}
\centering
\includegraphics[width=\linewidth]{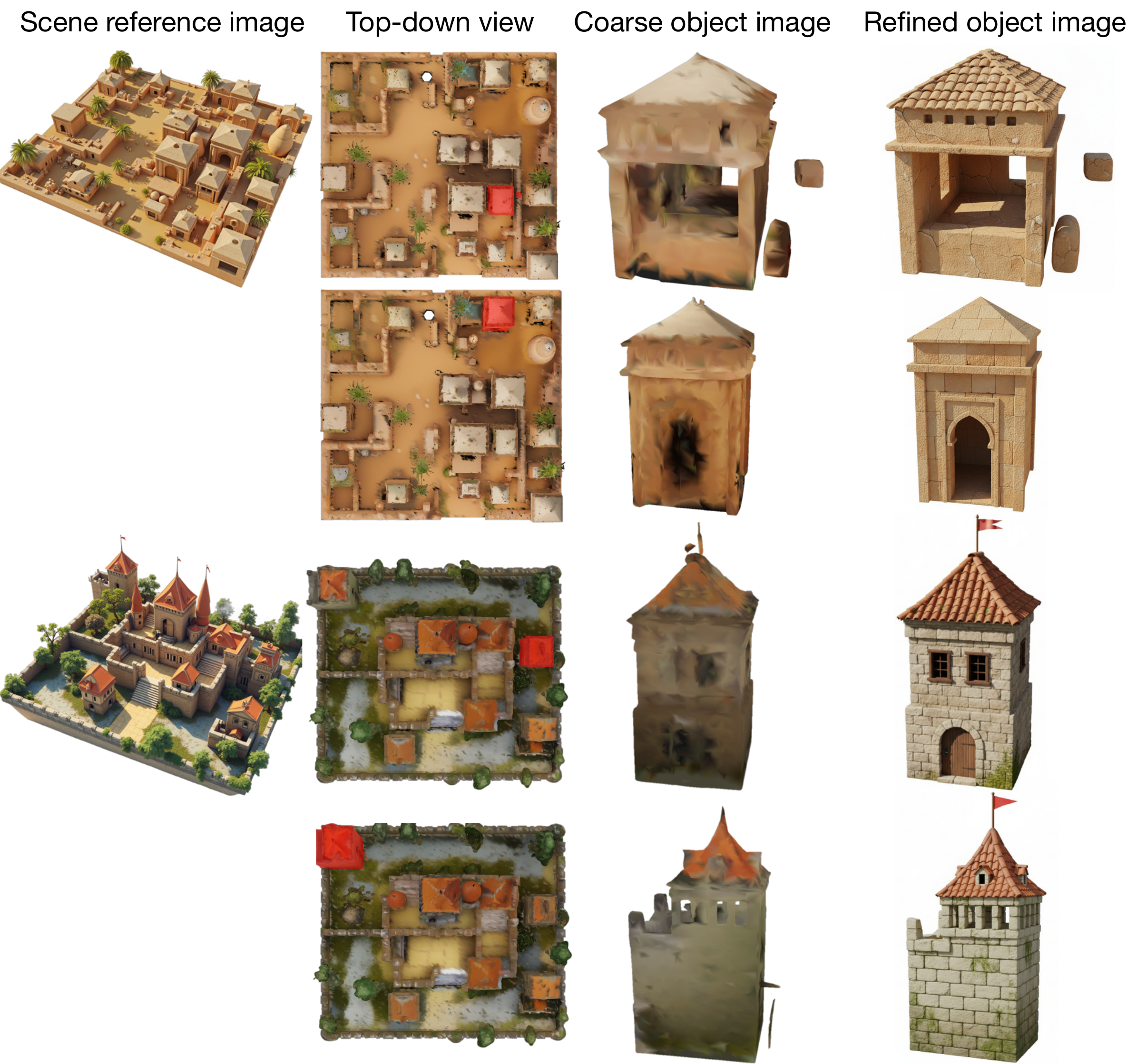}
\vspace{-0.6cm}
\caption{The reference scene image and the top-down render of the generated reconstruction, with the target object highlighted in red, are provided to the VLM to leverage scene context.
The VLM then refines the coarse object rendering to generate the final complete, detailed, and enhanced image of the object.}%
\label{fig:VLM_viz}
\vspace{-0.3cm}
\end{figure}

For training 3D shape enhancement in object-based scene generation, \name also requires conditioning images in addition to the input coarse shapes.
To obtain these images, we task a VLM with enhancing the rendered image of the coarse shape.
To provide enough context about the scene, we provide the VLM with an image of the entire scene, a top-down render of the scene with the object of interest highlighted, and a frontal render of the input coarse shape.
In \Cref{fig:VLM_viz}, we show visualizations of the inputs (from left to right: reference scene image, top-down render of the generated scene, and renders of the segmented coarse shapes), as well as the results from the VLM for the scenes in the main paper.
By providing the VLM with important information about object location, semantics, and surrounding context, we can generate disoccluded per-object images that are stylistically coherent with the overall scene.

\section{Additional qualitative results and comparisons}%
\label{supp sec: qualitative results}
\begin{figure*}[t!]
\centering
\includegraphics[width=\linewidth]{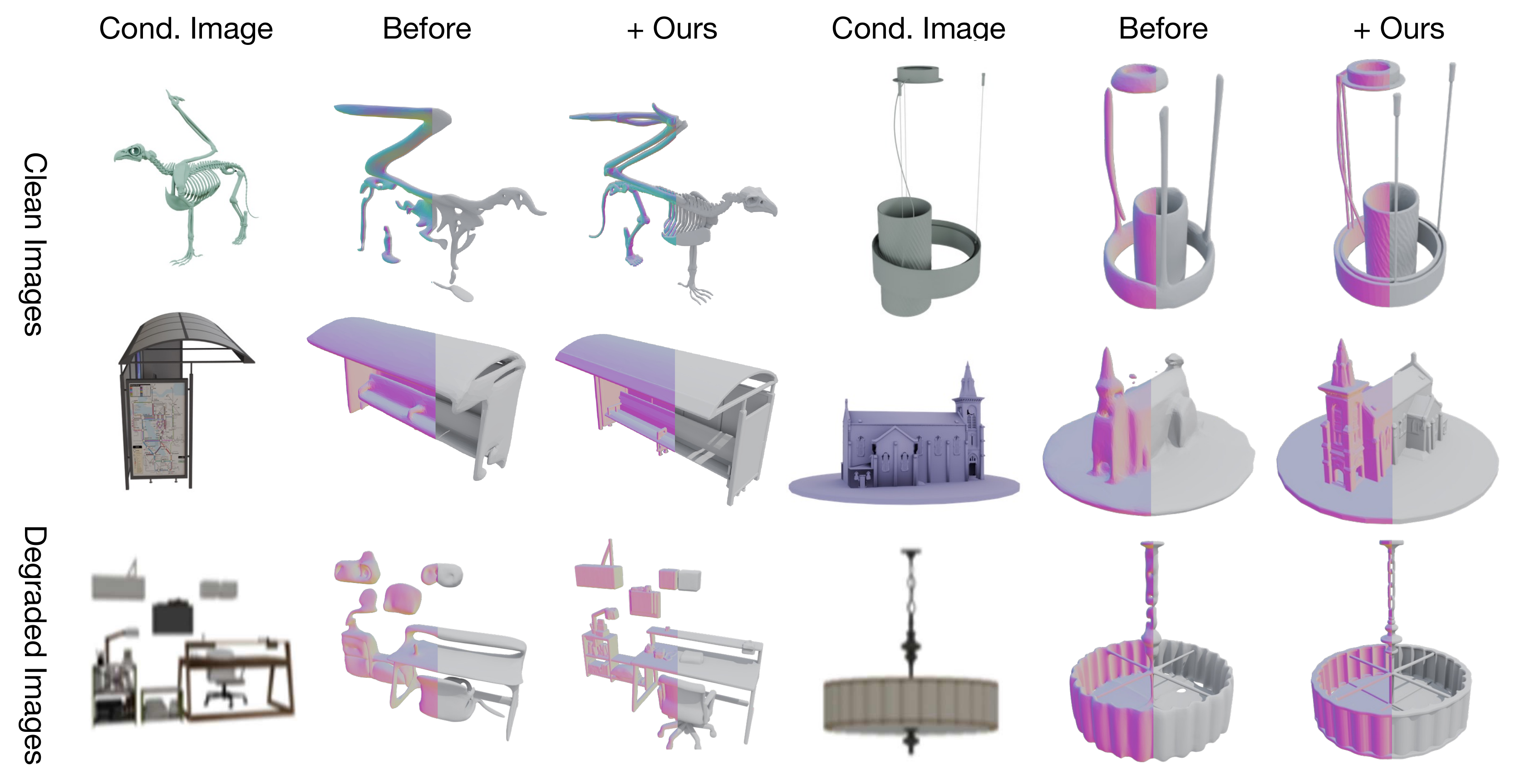}
\vspace{-0.5cm}
\caption{Qualitative results of 3D shape enhancement on the held-out validation set.
\name demonstrates its robustness, enhancing the shape even when both image and shape conditions are degraded.}%
\label{supp fig: enhance}
\vspace{-0.2cm}
\end{figure*}
\paragraph{Additional shape refinement results.}
\Cref{supp fig: enhance} provides additional qualitative results for 3D shape enhancement.
\name consistently reconstructs detailed and complete geometry from coarse inputs, even when conditioned on blurry or low-quality images, potentially due to the robust image augmentations used during training.

\begin{figure}
\centering
\includegraphics[width=0.6\linewidth]{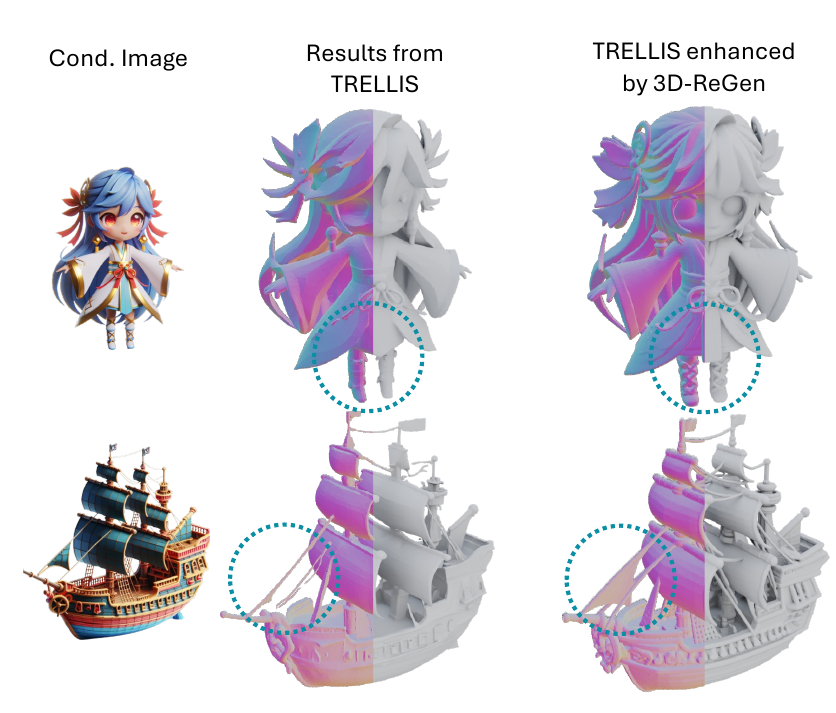}
\caption{Qualitative results for \name on 3D asset enhancement. \name shows promising improvements in recovering fine details and refining artifacts in the original generated shapes, resulting in outputs that better match the input image.}%
\label{fig:trellis_refine}
\end{figure}

\paragraph{Refining lower-quality 3D assets}
We present preliminary results showing that \name can enhance low-quality 3D assets arising from noisy or low-resolution scans, compression artifacts, legacy content, degraded scene objects, or, in this case, outputs from weaker generators. In \cref{fig:trellis_refine}, we use demo images and 3D assets generated by TRELLIS\footnote{\url{https://microsoft.github.io/TRELLIS/}} as input to \name, and demonstrate that \name can potentially improve the quality of existing 3D assets.

\begin{figure}[t!]
\centering
\includegraphics[width=0.9\linewidth]{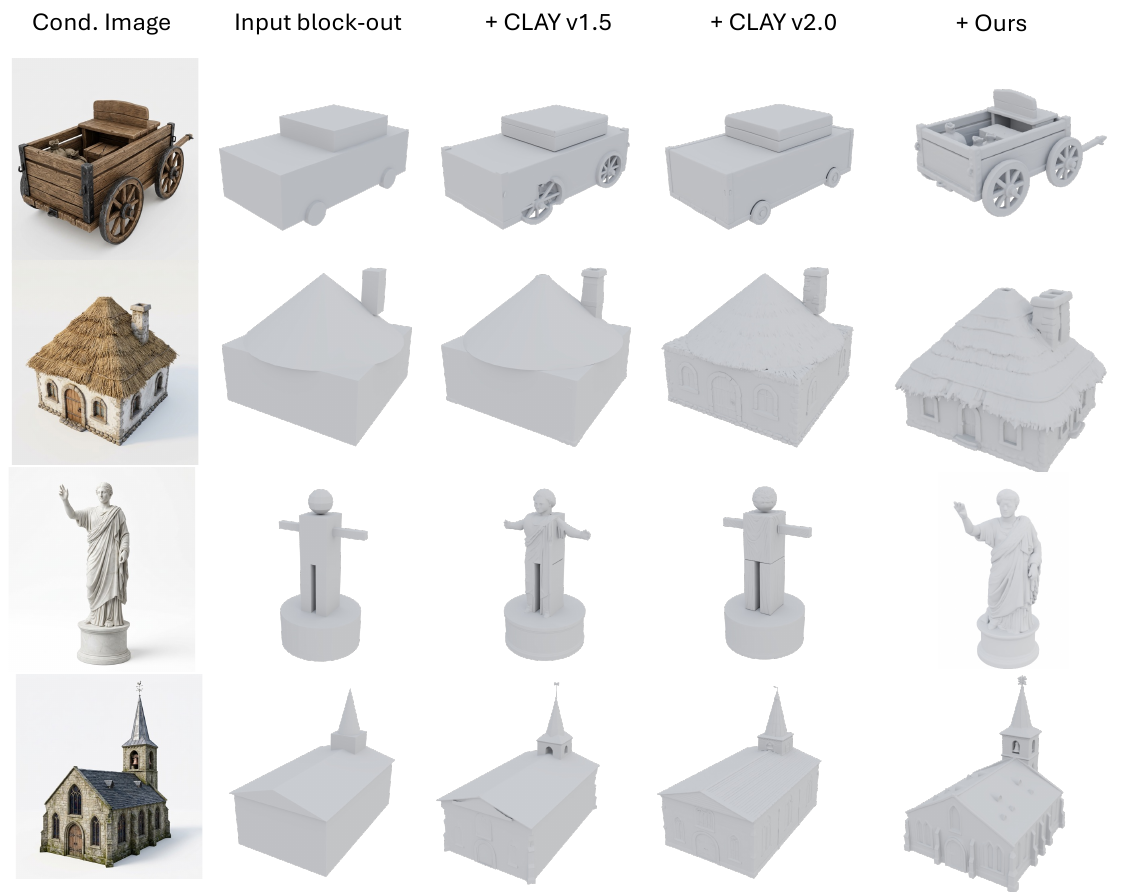}
\vspace{-0.4cm}
\caption{Qualitative results of \name on 3D shape regeneration from block-outs. Compared to CLAY, \name is capable of adding fine details while preserving the underlying shapes and proportions of the block-outs.}%
\label{fig:blockout}
\vspace{-0.6cm}
\end{figure}

\paragraph{3D shape generation from block-outs}
In a typical 3D creation workflow, artists start from a 2D concept image and a coarse 3D block-out that defines the overall extent, proportions, and rough geometry. We show that 3D \textbf{re}generation with \name naturally supports this setting: given simple block-out shapes and a guide image (\cref{fig:blockout}), \name synthesizes fine geometric details consistent with the image while preserving the block-out's structure and proportions. We compare against CLAY~\cite{zhang24clay:}, a point-cloud-conditioned 3D generator, using its official interface and point-cloud sampling tool to condition on the block-outs\footnote{\href{https://hyper3d.ai/}{Rodin.ai (Hyper3D): https://hyper3d.ai/}}. CLAY struggles to enrich the block-outs with fine-grained detail, highlighting the effectiveness of our 3D \textbf{re}generation formulation and curated paired training data even for extremely coarse shape guidance, without needing a dedicated architecture module or paired dataset generation for this task.

\begin{figure}[htbp]
\centering
\includegraphics[width=\linewidth]{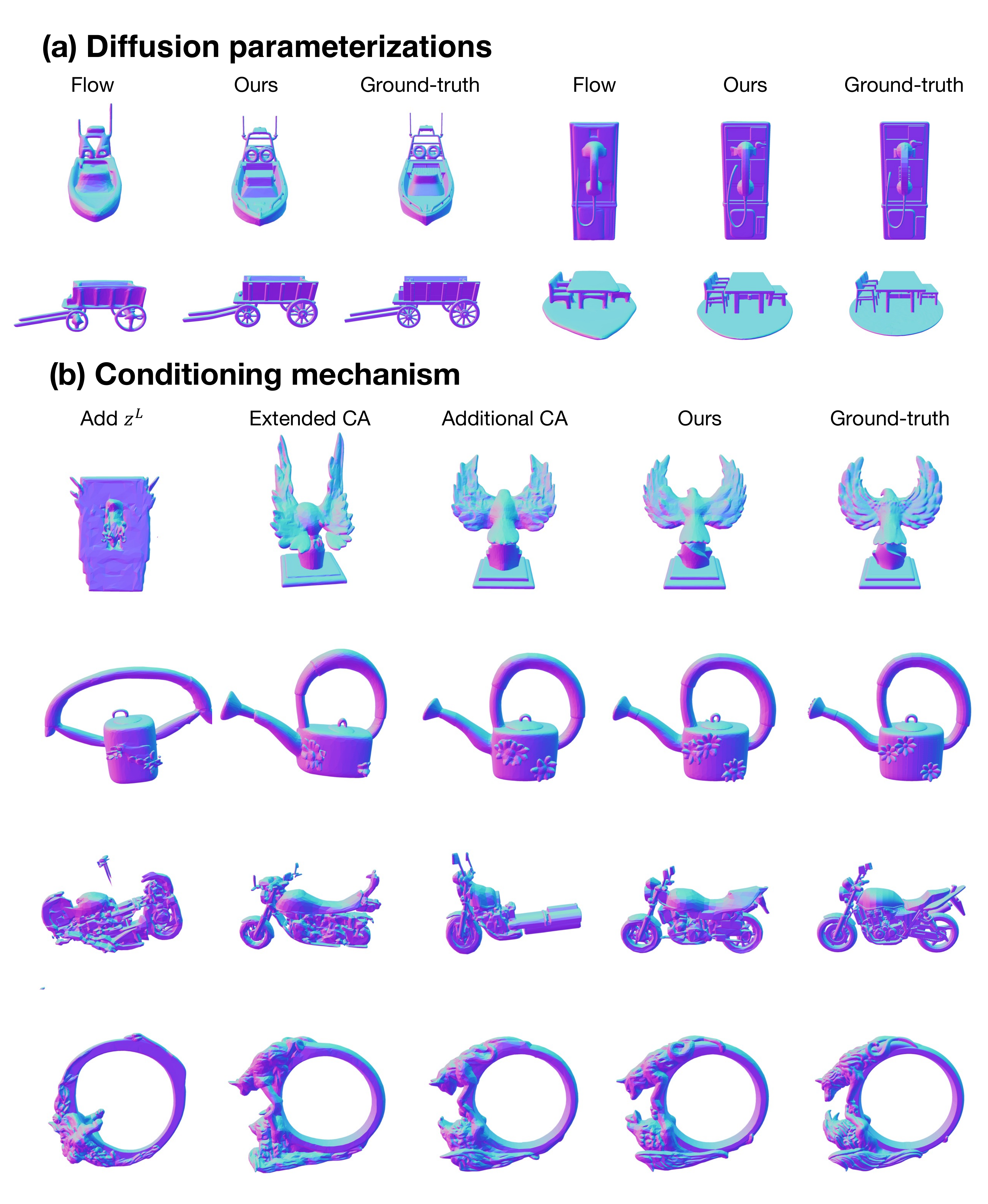}
\caption{More qualitative examples from the ablation study.
(\textbf{a}) Samples from the ablation study of different diffusion parameterizations in ~\cref{table: ablation_diffusion}.
(\textbf{b}) Samples from the ablation study of the conditioning mechanism in ~\cref{table: ablation_condition}.
The proposed method consistently outperforms other design choices with fine-grained details.}%
\label{supp fig: ablation}
\end{figure}

\paragraph{Ablation study.}
\Cref{supp fig: ablation} shows more qualitative samples from the ablation experiments reported in~\cref{table: ablation_condition} and~\cref{table: ablation_diffusion}.
To make our design choices feasible to evaluate, all ablation studies are conducted following the settings of early training stages in 3D enhancement tasks: models in~\cref{table: ablation_condition} use $C=512$ and train for 25K steps on $3\times3$ degraded shapes, while those in~\cref{table: ablation_diffusion} use $C=512$ and train for 25K steps on $2\times2$ degraded shapes.

\name produces sharper, more faithful geometric details than both (\textbf{a}) direct rectified-flow mappings from coarse to clean latents, and (\textbf{b}) alternative conditioning schemes such as cross-attention or additive latent injection. Crucially, our conditioning preserves the original diffusion parameterization and sampling dynamics; unlike additive perturbations or flow-based transports, it enables stable training while fully leveraging the capacity of the pretrained model.

\begin{table}[t!]
\caption{3D object reconstruction on GSO~\cite{downs22google} compared with Hunyuan3D-Omni~\cite{hunyuan3d2025hunyuan3d}.}%
\label{supp table: gso}
\centering
\resizebox{\linewidth}{!}{
\begin{tabular}{@{}cccccccc@{}}
\toprule
Model & \# View & CD ($\downarrow$) & F-score ($\uparrow$) & IoU ($\uparrow$) & PSNR ($\uparrow$) & SSIM ($\uparrow$) & LPIPS ($\downarrow$) \\ \midrule
Hunyuan3D-Omni~\cite{hunyuan3d2025hunyuan3d} & 1 & 0.0173 & 0.3267  & 0.5754  & 21.1630 & 0.9188 & 0.1280 \\ \midrule
\cellcolor{our_green} \name  & \cellcolor{our_green} 1  & \cellcolor{our_green} {\textbf{0.0135}}  & \cellcolor{our_green} {\textbf{0.3658}}  & \cellcolor{our_green} {\textbf{0.6438}}  & \cellcolor{our_green} {\textbf{21.8982}}  & \cellcolor{our_green} {\textbf{0.9315}}  & \cellcolor{our_green} {\textbf{0.1097}} \\ \midrule
Hunyuan3D-Omni~\cite{hunyuan3d2025hunyuan3d} & 2 & 0.0141 & 0.3812  & 0.6379  & 21.6921 & 0.9191 & 0.1246 \\ \midrule
\cellcolor{our_green} {\name}  & \cellcolor{our_green} 2  & \cellcolor{our_green} {\textbf{0.0112}}  & \cellcolor{our_green} {\textbf{0.4299}} & \cellcolor{our_green} {\textbf{0.7053}} & \cellcolor{our_green} {\textbf{22.9795}} & \cellcolor{our_green} {\textbf{0.9353}}  & \cellcolor{our_green} {\textbf{0.0985}}  \\ \midrule
Hunyuan3D-Omni~\cite{hunyuan3d2025hunyuan3d} & 3 & 0.0138 & 0.4190  & 0.6468  & 22.0323 & 0.9213 & 0.1221 \\ \midrule
\cellcolor{our_green} {\name}  & \cellcolor{our_green} 3 & \cellcolor{our_green} {\textbf{0.0090}}  & \cellcolor{our_green} {\textbf{0.4650}} & \cellcolor{our_green} {\textbf{0.7372}} & \cellcolor{our_green} {\textbf{23.6394}}  & \cellcolor{our_green} {\textbf{0.9385}}  & \cellcolor{our_green} {\textbf{0.0918}} \\ \midrule
Hunyuan3D-Omni~\cite{hunyuan3d2025hunyuan3d} & 4 & 0.0136 & 0.4248  & 0.6603  & 22.0141 & 0.9220 & 0.1203 \\ \midrule
\cellcolor{our_green} {\name}  & \cellcolor{our_green} 4 & \cellcolor{our_green} \textbf{0.0081} & \cellcolor{our_green} \textbf{0.4913} & \cellcolor{our_green} \textbf{0.7574}  & \cellcolor{our_green} \textbf{24.2754}  & \cellcolor{our_green} \textbf{0.9408}  & \cellcolor{our_green} \textbf{0.0873} \\
\bottomrule
\end{tabular}
}
\vspace{-0.3cm}
\end{table}
\paragraph{Comparisons with CLAY~\cite{zhang24clay:}.}%
\label{supp sec: CLAY}
\begin{figure}[t!]
\centering
\includegraphics[width=\linewidth]{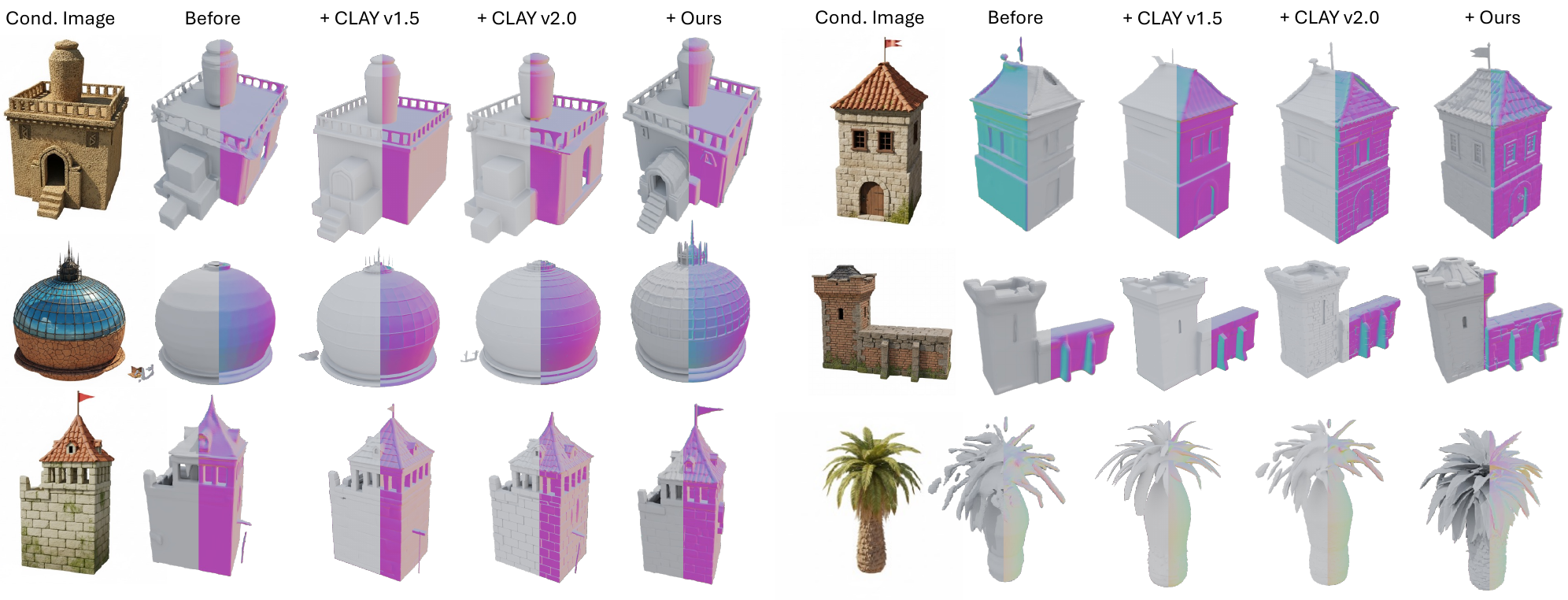}
\vspace{-0.4cm}
\caption{
Additional qualitative comparisons against 3D-enhancement baselines.
We obtain coarse 3D conditions by extracting parts from the original scene~\cite{li25sparc3d:}.
Leveraging coarse 3D shapes and guide images, \name regenerates fine-grained 3D shape details and outperforms the state-of-the-art point-cloud-based method for conditional 3D generation, CLAY~\cite{zhang24clay:}.
}%
\label{fig:clay_enhancement}
\vspace{-0.6cm}
\end{figure}
We provide additional qualitative comparisons with CLAY~\cite{zhang24clay:}, a point-cloud and image-conditioned 3D shape generator, on the task of 3D shape enhancement. We include results from both CLAY v1.5 and v2.0 obtained via the official interface\footnote{\href{https://hyper3d.ai/}{Rodin.ai (Hyper3D): https://hyper3d.ai/}}. We feed the coarse shapes to CLAY's point-cloud sampling tools, together with the guide image, as conditioning inputs.
As shown in \cref{fig:clay_enhancement}, both versions of CLAY often fail to recover fine details specified by the guide image, and may leave broken geometry unresolved (right column, last row) or introduce floaters. These results highlight the benefits of our unified architectural design choices and dataset curation process for 3D \textbf{re}generation in shape enhancement.
\paragraph{Comparisons with Hunyuan3D-Omni~\cite{hunyuan3d2025hunyuan3d}.}%
\label{supp sec: hunyuan3d}

We further compare our method with Hunyuan3D-Omni~\cite{hunyuan3d2025hunyuan3d}, a recent \emph{concurrent} preprint.
The two works differ significantly in scope and mechanisms, although both can be used to improve the controllability of 3D diffusion models.

(\textbf{a}) \textbf{Task scope.}
Hunyuan3D\-Omni focuses on improving controllability within the realm of 3D \emph{generation}.
In contrast, we introduce a novel and unified 3D \emph{regeneration} framework that handles a broad set of downstream tasks, including enhancement, reconstruction, and editing, none of which are addressed by Hunyuan3D-Omni.

(\textbf{b}) \textbf{Data construction.}
We present several pretext tasks and data pipelines that automatically transform base 3D datasets into task-specific triplets without extra annotations.
In contrast, Hunyuan3D-Omni primarily uses base datasets as-is or relies on external sources and annotations~\cite{yan2025posemaster}.

(\textbf{c}) \textbf{Architecture.}
Hunyuan3D-Omni converts all conditioning signals into point clouds (up to 2048 points) and encodes them using a shallow MLP as a point encoder for cross-attention conditioning. 
While we also use a point-cloud representation, our method encodes it with a \vecset VAE, producing compact latent tokens that capture both the global structure and dense geometric detail, potentially beneficial for tasks requiring precise geometry preservation, such as 3D shape enhancement.

\begin{figure}[t!]
\centering
\includegraphics[width=\linewidth]{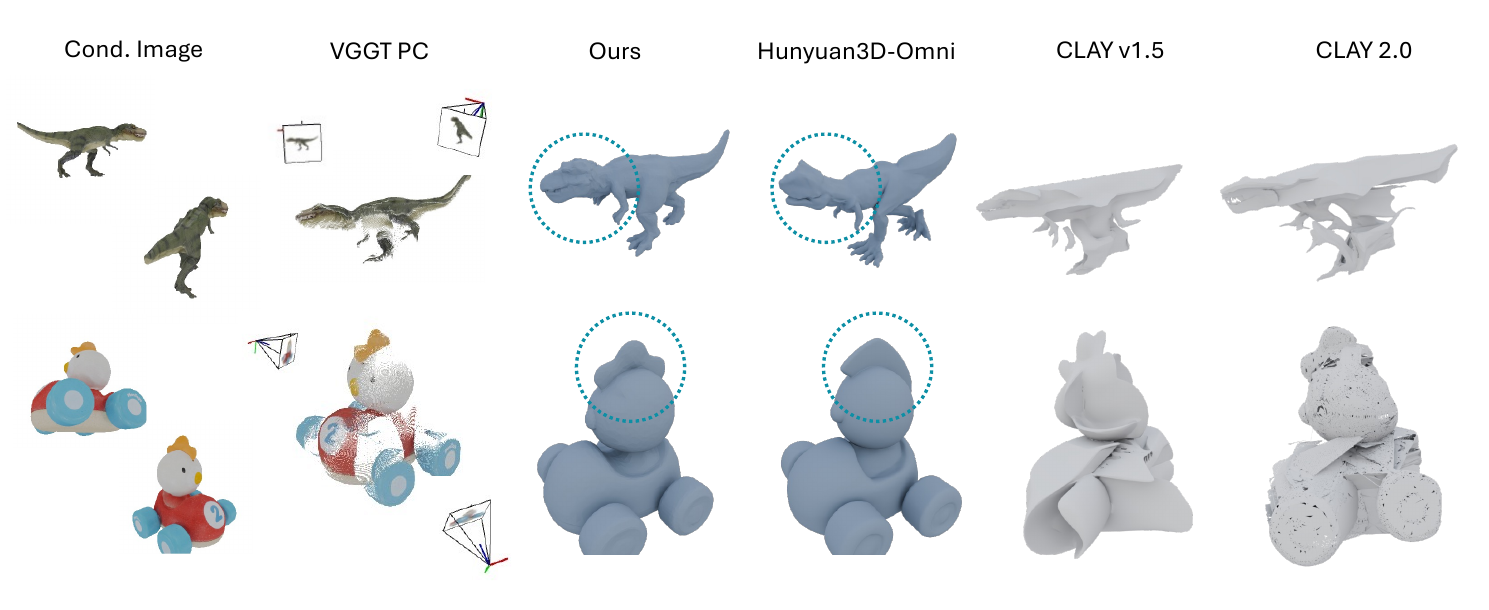}
\vspace{-0.5cm}
\caption{Qualitative comparison of two-view object reconstruction with Hunyuan3D-Omni ~\cite{hunyuan3d2025hunyuan3d} and  CLAY ~\cite{zhang24clay:}.
\name produces more complete, faithful and higher-fidelity shapes.}%
\vspace{-0.3cm}
\label{supp fig: hunyuan}
\end{figure}

To compare these two models, we evaluate 3D object reconstruction on the GSO dataset~\cite{downs22google} when varying the number of conditioning views.
Given an initial point-map prediction, Hunyuan3D-Omni subsamples 2048 points and encodes them using a shallow point encoder, whereas \name encodes the same input into \vecset tokens via a pretrained 3D VAE\@.
As shown in~\cref{supp table: gso} and~\cref{supp fig: hunyuan}, \name achieves better reconstruction across view counts.

In~\cref{supp fig: hunyuan}, we also include qualitative results generated by CLAY \cite{zhang24clay:}.
Unlike Hunyuan3D-Omni, which avoids per-modality cross-attention heads, CLAY injects 3D controls as separate, trainable residual cross-attention branches in the DiT (with tunable $\alpha_i$ weights).
We include both CLAY v1.5 and v2.0 results using the official website\footnote{\href{https://hyper3d.ai/}{Rodin.ai (Hyper3D): https://hyper3d.ai/}}. As shown in ~\cref{supp fig: hunyuan}, CLAY struggles to recover full 3D shapes when input point clouds are incomplete.
Since CLAY and Hunyuan3D-Omni condition the diffusion model in fundamentally different spaces, they could potentially be combined to further improve controllability in future 3D re-generation systems.

\section{More Related Works}%
\label{supp sec: related works}

\paragraph{3D Downstream Tasks.}
Going beyond 3D generation, we present \name as a unified framework for 3D geometry regeneration, built upon an image-to-3D generative model that directly produces SDF representations.
Leveraging a versatile control space and self-supervision with base 3D datasets, our approach enables a controllable 3D generation pipeline supporting diverse downstream tasks, including 3D enhancement, shape reconstruction, and editing. This application enables us to leverage key advantages of 3D diffusion models in downstream tasks: (1) clean and coherent surface extraction via SDFs, (2) a 3D-native latent space, and (3) strong generative capacity.

For \textit{3D enhancement}, existing methods~\cite{zhang2025supercarve, deng24detailgen3d:, li2024craftsman3d, ye2024stablenormal} enhance geometry via multi-view normal refinement but inherit view inconsistency from multi-view diffusion models.
In contrast, \name directly refines \vecset latents using paired coarse-fine mesh data, ensuring coherent, view-consistent geometry.

For \textit{3D reconstruction}, large reconstruction models~\cite{hong24lrm:, zhang24gs-lrm:, tang24lgm:, wei24meshlrm:, jin25lvsm:} generate radiance fields from limited views but are typically deterministic, producing blurry or inconsistent shapes in unseen regions.
\name serves as a generative 3D enhancer that adaptively balances observed evidence and learned 3D priors, enabling robust reconstruction across varying view settings.

For \textit{3D editing,} prior approaches rely on expensive inversion~\cite{li2025voxhammer}, optimization~\cite{liu2024make, sella2023vox, zhuang2024tip}, or 2D-to-3D lifting~\cite{chen2024generic, barda2025instant3dit, chen24dge}, often leading to view inconsistency and shape distortion.
In contrast, \name performs feed-forward editing directly in the 3D \vecset space, preserving geometric consistency and faithfully retaining unedited regions.

\end{document}